\documentclass[10pt,twocolumn,letterpaper]{article}

\usepackage[pagenumbers]{cvpr} 

\usepackage{graphicx}
\usepackage{amsmath}
\usepackage{amssymb}
\usepackage{booktabs}

\usepackage[pagebackref,breaklinks,colorlinks]{hyperref}

\usepackage[capitalize]{cleveref}
\crefname{section}{Sec.}{Secs.}
\Crefname{section}{Section}{Sections}
\Crefname{table}{Table}{Tables}
\crefname{table}{Tab.}{Tabs.}

\newcommand*{\affaddr}[1]{#1} 
\newcommand*{\affmark}[1][*]{\textsuperscript{#1}}

\usepackage[symbol]{footmisc}

\begin{document}




\newcommand{\hlc}[2][color]{{\sethlcolor{#1}\hl{#2}}}

\newif\ifdraft
\drafttrue

\ifdraft
\definecolor{darkg}{rgb}{0.0,0.5,0.0}
\newcommand{\dcc}[1]{{\color{red}[\textbf{Danny:} #1]}}
\newcommand{\kac}[1]{{\color{blue}[\textbf{Kfir:} #1]}}
\newcommand{\avc}[1]{{\color{darkg}[\textbf{Andrey:} #1]}}

\newcommand{\dc}[1]{{\color{red}#1}}
\newcommand{\ka}[1]{{\color{blue}#1}}
\newcommand{\av}[1]{{\color{darkg}#1}}



\newcommand{\rev}[1]{{\color{brown}#1}}

\newcommand{\drop}[1]{}



\newcommand{\cradd}[1]{{\color{black}#1}}
\newcommand{\crmv}[1]{{\color{black}#1}}

\else
\newcommand{\dcc}[1]{}
\newcommand{\kac}[1]{}
\newcommand{\avc}[1]{}
\newcommand{\dc}[1]{{\color{black}#1}}
\newcommand{\ka}[1]{{\color{black}#1}}
\newcommand{\av}[1]{{\color{black}#1}}
\fi

\newcommand{\bbf}{{\bf F}}
\newcommand{\bbi}{{\bf I}}
\newcommand{\Mi}{\hat{M}_{l-1}}
\newcommand{\Mo}{\hat{M}_{l}}
\newcommand{\dVi}{\Delta \hat{V}_{l-1}}
\newcommand{\dVo}{\Delta \hat{V}_{l}}
\newcommand{\Vi}{\hat{V}_{l-1}}
\newcommand{\Vo}{\hat{V}_{l}}
\newcommand{\dEo}{\Delta \hat{E}_{l}}
\newcommand{\Ci}{\hat{C}_{l-1}}
\newcommand{\Wo}{W_{l}}
\newcommand{\partmesh}{PartMesh}
\newcommand{\eps}{\varepsilon}
\newcommand{\norm}[1]{\left\Vert #1 \right\Vert_2}
\newcommand{\localalpha}{\alpha}
\newcommand{\localalphaf}{B}

\def\naive{na\"{\i}ve\xspace}
\def\Naive{Na\"{\i}ve\xspace}

\newcommand{\w}{$\mathcal{W}$\xspace}
\newcommand{\wplus}{$\mathcal{W}+$\xspace}

\makeatletter
\DeclareRobustCommand\onedot{\futurelet\@let@token\@onedot}
\def\@onedot{\ifx\@let@token.\else.\null\fi\xspace}

\def\eg{\emph{e.g}\onedot}
\def\Eg{\emph{E.g}\onedot}
\def\ie{\emph{i.e}\onedot}
\def\Ie{\emph{I.e}\onedot}
\def\etc{\emph{etc}\onedot}
\def\etal{\emph{et al}\onedot}
\makeatother

\raggedbottom

\makeatletter
\def\blfootnote{\xdef\@thefnmark{}\@footnotetext}
\makeatother

\newcommand{\ccbync}{\href{https://creativecommons.org/licenses/by-nc/4.0/legalcode}{CC BY-NC 4.0}}

\newcommand{\cczero}{\href{https://creativecommons.org/publicdomain/zero/1.0/}{CC0 1.0}}

\newcommand{\ccbyncsa}{\href{https://creativecommons.org/licenses/by-nc-sa/4.0/}{CC BY-NC-SA 4.0}}

\newcommand{\nvsrc}{\href{https://nvlabs.github.io/stylegan2/license.html
}{Nvidia Source Code License-NC}}

\newcommand{\bsd}{\href{https://opensource.org/licenses/BSD-3-Clause}{BSD 3-Clause}}

\newcommand{\mitlic}{\href{https://opensource.org/licenses/MIT}{MIT License}}

\newcommand{\adblic}{\href{https://github.com/utkarshojha/few-shot-gan-adaptation/blob/main/LICENSE.txt}{Adobe Research License}}


\newcommand{\naturals}{\mathbb{N}}
\newcommand{\reals}{\mathbb{R}}
\newcommand{\attmask}{M}
\newcommand{\pixelmod}{i}
\newcommand{\textmod}{p}
\newcommand{\textemb}{e}
\newif\ifwatermark
\watermarktrue
\draftfalse

\title{Sketch-Guided Text-to-Image Diffusion Models}

\author{%
Andrey Voynov \affmark[1], Kfir Aberman\affmark[1], Daniel Cohen-Or\footref{note}~~\affmark[1,]\affmark[2]\\
\small{\affaddr{\affmark[1]Google Research,~~}}\small{\affaddr{\affmark[2]The Blavatnik School of Computer Science, Tel Aviv University}}\
}


\twocolumn[{
\renewcommand\twocolumn[1][]{#1}
\maketitle
\begin{center}
  \vspace{-14pt}
  \includegraphics[width=0.999\textwidth]{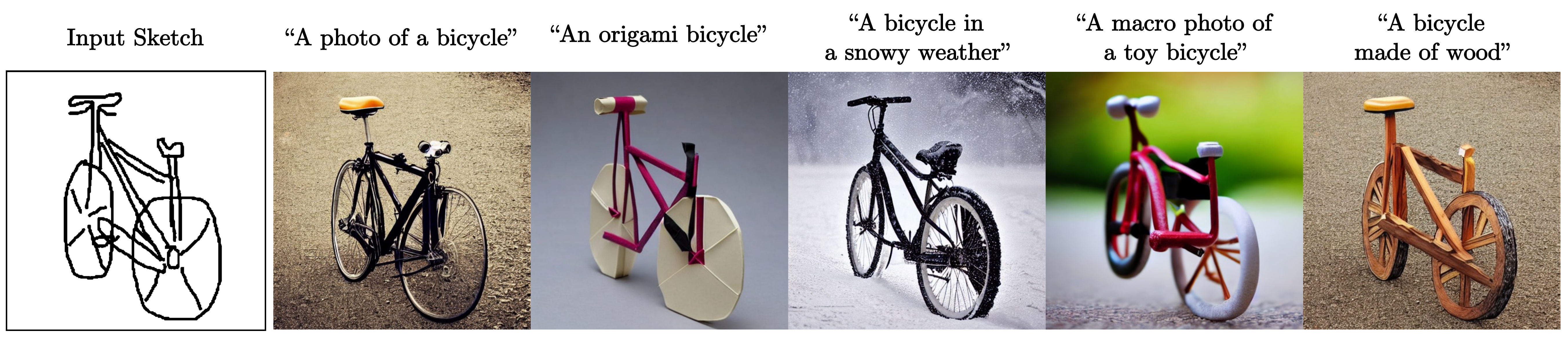}
  \vspace{-12pt}
  \captionof{figure}{Given a sketch and a text-prompt, our method uses the sketch to guide a pretrained text-to-image diffusion model during inference time. The method allows producing diverse results that correspond to the text-prompt and follow the spatial layout of the sketch.}
  \label{fig:teaser}
\end{center}
}]

\maketitle
\begin{abstract}

Text-to-Image models have introduced a remarkable leap in the evolution of machine learning, demonstrating high-quality synthesis of images from a given text-prompt. However, these powerful pretrained models still lack control handles that can guide spatial properties of the synthesized images. 
In this work, we introduce a universal approach to guide a pretrained text-to-image diffusion model, with a spatial map from another domain (e.g., sketch) during inference time. Unlike previous works, our method does not require to train a dedicated model or a specialized encoder for the task.

Our key idea is to train a Latent Guidance Predictor (LGP) - a small, per-pixel, Multi-Layer Perceptron (MLP) that maps latent features of noisy images to spatial maps, where the deep features are extracted from the core Denoising Diffusion Probabilistic Model (DDPM) network.
The LGP is trained only on a few thousand images and constitutes a differential guiding map predictor, over which the loss is computed and propagated back to push the intermediate images to agree with the spatial map.
The per-pixel training offers flexibility and locality which allows the technique to perform well on out-of-domain sketches, including free-hand style drawings.
We take a particular focus on the sketch-to-image translation task, revealing a robust and expressive way to generate images that follow the guidance of a sketch of arbitrary style or domain.

\end{abstract}

\footnotetext[2]{\label{note}Performed this work while working at Google.}

\section{Introduction}

Large text-to-image diffusion models \cite{ramesh2022hierarchical,saharia2022photorealistic,rombach2021highresolution} have been an inspiring tool for content creation and editing, enabling synthesis of diverse images with unprecedented quality that follow a given text-prompt.
Despite the semantic guidance provided by the text-prompt, these models still lack intuitive control handles that can guide spatial properties of the synthesized images. In particular, guiding a pertained text-to-image diffusion model during inference with a spatial map from another domain, such as sketch, is yet an open challenge.

A possible attempt is by training a dedicated encoder to map the guiding image into the
latent space of the pretrained unconditional diffusion model ~\cite{PITI22}. The trained encoder, however, performs well in-domain, but struggles with out-of-domain free-hand sketches.

In this work, we introduce a generic approach to guide the inference process of a pretrained text-to-image diffusion model with a spatial map. Our key idea is to use a small multi-layer perceptron (MLP) network that is trained to map latent features of noisy images to spatial maps, where the latent features are extracted from the core network of the diffusion model. The trained MLP serves as a latent guidance predictor, over which the loss with a target spatial map is computed and propagated back to push the intermediate image to agree with the map.

The latent guidance predictor is trained in a self-supervised fashion and learns to translate features of images with different noise levels into encoded spatial maps, where the noise scheduling corresponds to the noise scheduling of the diffusion process. Importantly, the latent guidance predictor is trained, and operates independently, on each latent pixel in the latent space, rather than on the whole image. Hence, it is sufficient to train it with a few thousand images only, which is a few orders of magnitude less than the required amount to train a dedicated image-to-image translation model. Yet, the latent guidance predictor is generic and domain-oblivious, in the sense that it can operate on out-of-domain guiding maps. Hence, our method can accept free-hand sketches inputs, as in Figure \ref{fig:teaser}, and generate diverse results that correspond to the text-prompt and follow the spatial layout of the sketch.

In our experiments, we demonstrate sketch-guided text-to-image synthesis results on various domains, including free-hand style drawing. We conduct experiments and ablation studies to analyze the performance of various components of our method and present comparisons to other image translation approaches. In addition, we show that our general framework can be applied to other spatially guided text-to-image tasks such as saliency-guided inpainting and horizon control. Throughout our examples in the paper, we demonstrate that our method can be applied to a rich variety of sketch styles from diverse domains, which is the key advantage of our approach. Project page: \url{sketch-guided-diffusion.github.io}
\section{Related Work}
\label{sec:prior}

\subsection{Image-to-Image Translation}
Image-to-image translation has been a long-standing task in the computer vision domain with a myriad of explorations and prior works \cite{pang2021image}.
One common architecture to perform image translation is conditional generative adversarial networks \cite{isola2017image, zhu2017toward,wang2018high,park2019semantic,zhu2017unpaired}
that translate an input domain to an output domain with a discriminator to bridge the gap between real images and fake ones. Commonly, these methods tackle each task independently and use task-specific datasets and models. Considering the commonality between tasks, some research efforts \cite{zhang2021ufc,huang2022multimodal,kutuzova2021multimodal} aim to learn a unified model for diverse translation tasks via multi-task training. 

A special case of image-to-image translation is the task of sketch-to-photo task~\cite{xiang2022adversarial,ghosh2019interactive,gao2020sketchycoco}. SketchyGAN~\cite{chen2018sketchygan} uses edge-preserving image augmentations to train a Generative Adversarial Network (GAN), ContextualGAN~\cite{lu2018image} leverages conditional GAN with joint image-sketch representation. CoGS~\cite{ham2022cogs}
minimizes the distances between the embeddings of
the input sketch and the corresponding ground truth real
image in the vector-quantized space of a VQ-GAN~\cite{esser2021taming}. In contrast, our work leverages a pretrained text-to-image generative prior of general images and treats the image translation problems as downstream tasks.

\subsection{Diffusion Models}
DDPM introduced unprecedented quality of conditional and unconditional image synthesis. 
\cite{ho2020denoising,dhariwal2021diffusion,ho2022cascaded}, rivaling GAN-based methods both in visual quality and sampling diversity.
In particular, \cite{saharia2022palette} uses diffusion models to solve various image translation tasks, but tackles each task independently and trains a model from scratch for each task. 

More recently, diffusion models have demonstrated unprecedented quality for text-to-image synthesis and editing tasks, when large models are trained on pairs of text and images \cite{ramesh2022hierarchical,saharia2022photorealistic,rombach2021highresolution,ruiz2022dreambooth,hertz2022prompt}. Our approach builds on these key advances, and we show how a pretrained text-to-image diffusion model can be guided by a spatial map from a different domain and serve as a universal generative prior that facilitates various image translation tasks. Note that models like Make-a-Scene~\cite{gafni2022make} and eDiffi~\cite{balaji2022ediffi} allow the user a particular type of control by enabling them to provide a semantic segmentation map to control the composition of the elements in the synthesized image.

ILVR~\cite{choi2021ilvr} proposes to iteratively refine the diffusion process using a noisy reference image at each time-step during inference, enabling control of the amount of high-level
semantics being adapted from the images. In addition, SDEdit 
\cite{meng2021sdedit} suggests to add noise to the input guiding image, halfway of the forward diffusion process, then denoise it in a reverse process with a guiding text. Both approaches enable to guide the model with an image where the guiding image should lay in the RGB domain and the fidelity to the spatial property of the guiding image is limited and random.

A closely related approach is the recent work of Wang et al.~\cite{PITI22}  that suggests to use a pretrained unconditional diffusion model for various image translation tasks, by training a specialized, per-task, encoder~\cite{richardson2020encoding} to map spatial maps into the latent space of the diffusion model. While their approach requires dedicated large-scale training to train the encoder, we use a light weight training (only a few thousand images are required) of a small MLP which is trained per-pixel, and thus offers a generalization that extends beyond the domain defined by the training data.

\section{Method}
\label{sec:method}

In this section, we describe the main steps of the proposed spatially-guided text-to-image synthesis approach. Although our method is generic, for the ease of reading, in the method description we are focused on the sketch-to-image task, and in Section~\ref{sec:exp}, we show how the very same approach also works for different tasks. 

The key idea of our method is to guide the inference process of a pretrained text-to-image diffusion model with an edge predictor that operates on the internal activations of the core network of the diffusion model, encouraging the edge of the synthesized image to follow a reference sketch. Our edge predictor is an MLP network, that operates per-pixel, and is trained to map features of noisy images into spatial edge maps. The training procedure which is performed only one time, requires a few thousand images only, and takes only about an hour on a single GPU. 

\subsection{Latent Edge Predictor}

\begin{figure}
    \centering
    \includegraphics[width=\columnwidth]{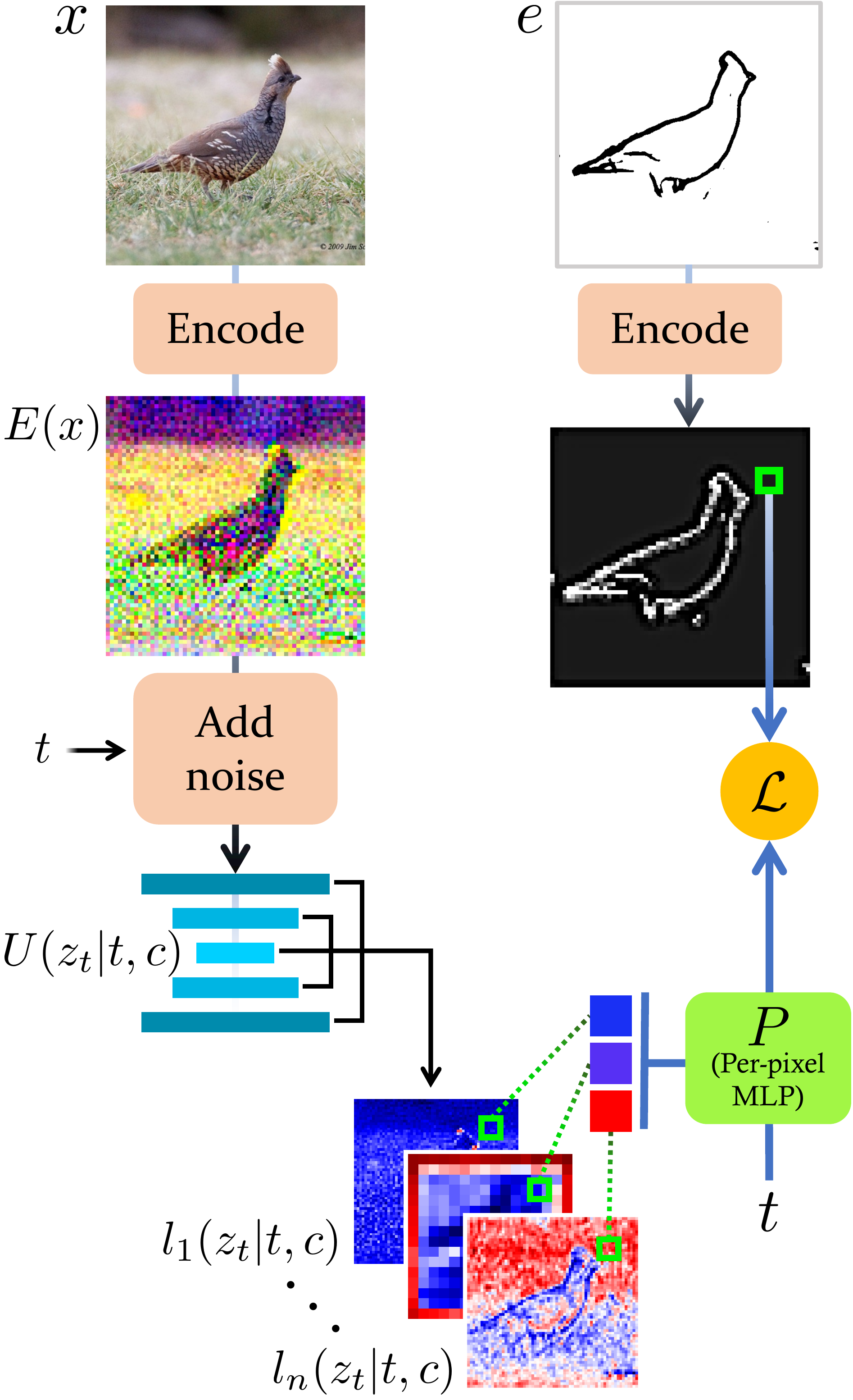}
    \caption{{\bf Training scheme of the Latent Edge Predictor.} Given an image $x$, we first encode it and add noise to get $z_t$, then pass $z_t$ through the core U-net network of a DDPM, and extract a set of latent features $l_1(z_t | t, c), \dots, l_n(z_t | t, c)$. Then our Latent Edge Predictor, which is a per-pixel MLP, is trained to map each pixel in the concatenated features to the corresponding pixel in the encoded edge map $e$.} 
    \label{fig:my_label}
\end{figure}

Our first goal is to train an MLP that guides the image generation process with a target edge map. The MLP is trained to map the internal activations of a denoising diffusion model network into spatial edge maps, as depicted in Figure~\ref{fig:my_label}. Inspired by \cite{baranchuk2021label}, we extract our activations from a fixed sequence of intermediate layers in the core U-net network $U$ of the diffusion model. Formally, for an input tensor $w$, we denote $\bbf(w|  c, t)= \left[l_1(w |  c, t), \dots, l_n(w|  c, t)\right]$ as the concatenated activations of selected internal layers $\{l_1, \dots, l_n\}$, when $w$ is processed by the network with a conditioning text-prompt $c$ and noise level $t$. Since activations from different layers may have different spatial resolution, we resize them to match the spatial dimensions of the input $w$ and concatenate them alongside the channel dimension. The input dimension of the MLP is then the sum of the number of channels of the selected activations.

Our training corpus $\mathcal{D}$ is formed by triplets $(x, e, c)$ of an image, edge map, and a corresponding text caption, respectively. Since our work is implemented with latent diffusion models (specifically Stable Diffusion), we use the model encoder $E$ to preprocess the images and the edge maps. In order to encode the edge map, we convert it into a 3-channel image by replicating its intensity channel.
Thus, in practice the input tensor is the encoded image with additive Gaussian noise, $z_t = \alpha_t \cdot E(x) + \mu_t \cdot \xi $, where $0\leq\alpha_t, \mu_t\leq 1$ are the blending scalars that is dictated by the noise scheduling of the diffusion model, and the MLP is trained to map the concatenated features $\bbf(z_t | c, t)$ to the encoded edge map $E(e)$.

In order to consider the noise level of the input, the MLP also receives $t$ and its positional encoding as $\sin(2\pi t \cdot 2^{-l}),\ l = 0, \dots, 9$.
The output dimension of the MLP is equal to the number of output channels of $E$ (4 in the case of Stable Diffusion). Each spatial position $(i, j)$, of the latent pixel $\bbf(z | c, t)_{ij}$ is translated to the corresponding latent edge $E(e)_{ij}$ by $P$, thus, the training objective of our latent edge predictor $P$ is
\begin{equation}
    \mathcal{L} = \mathop{\mathbb{E}}\limits_{(x, e, c) \sim \mathcal{D}} \mathop{\mathbb{E}}\limits_{\substack{t \sim \mathcal{U}([0, 1]) \\ \xi \sim \mathcal{N}(0, 1)}} \sum_{i,j}\|P(\mathbf{F}(z_t|c, t)_{i, j}, t)-E(e)_{ij}\|^2,
\label{eq:loss_main}
\end{equation}
where $P$ is applied to each latent pixel independently.

Once optimized with the objective $\mathcal{L}$, the model $P$ constitutes a per-spatial location differential predictor of encoded edges for an encoded image with noise level $t$. Due to the per-pixel nature of the architecture, the MLP is trained to predict edges in a local manner,  being agnostic to the domain of the image. In addition, it enables training on a relatively small corpus (a few thousand images), in reasonable training time (One hour on a single A100 GPU).

We next show how such a component can serve as a guidance through the diffusion process. 

\begin{figure}
    \centering
    \includegraphics[width=\columnwidth]{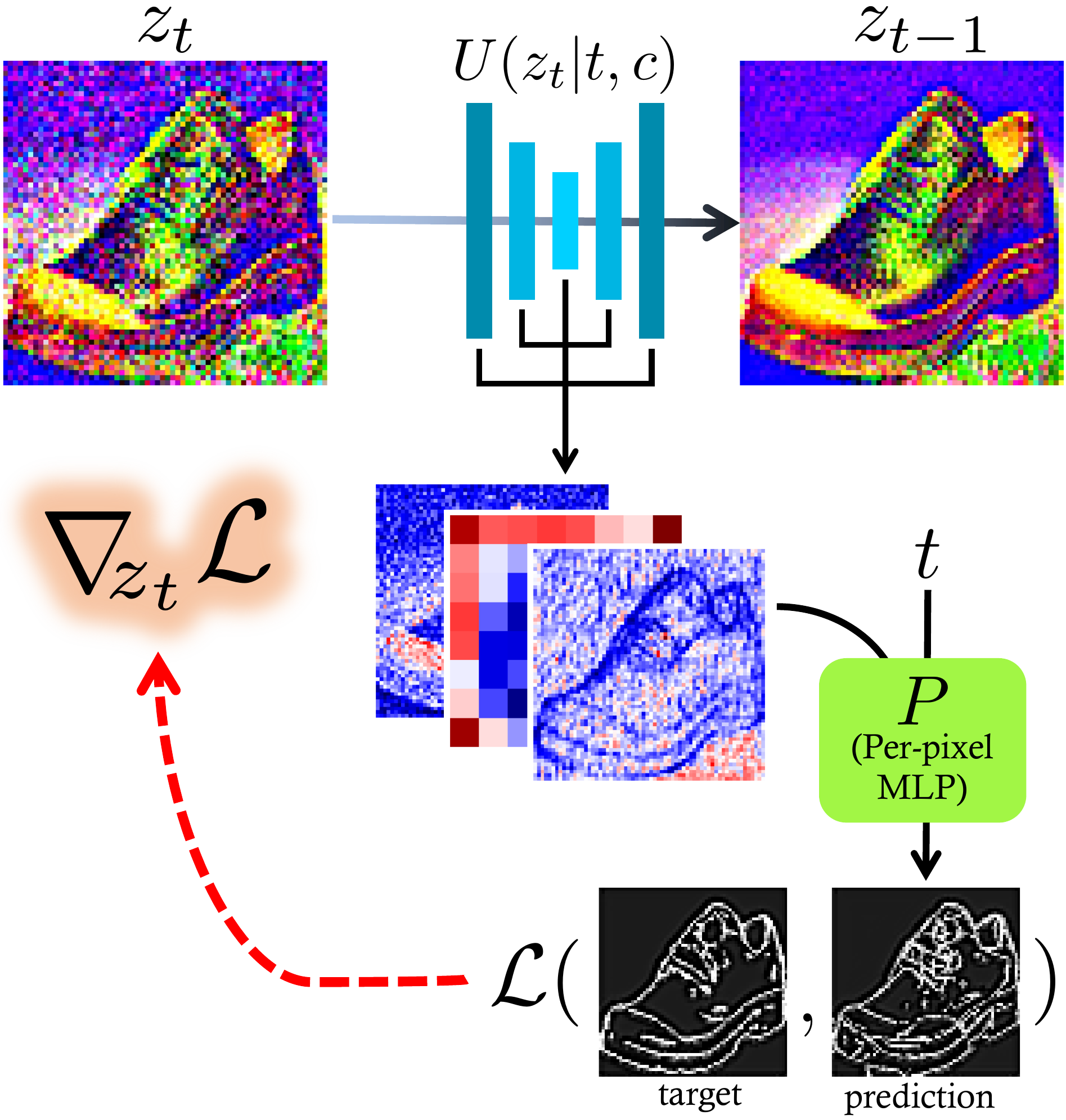}
    \caption{{\bf Sketch-Guided Text-to-Image Synthesis Scheme.} Given an encoded noisy image $z_t$, our method extracts its deep features during the inference process of a text-to-image diffusion model, conditioned on a caption $c$. In each of the denoising steps $t$, we aggregate the intermediate model features, and pass them in our per-pixel Latent Edge Predictor $P$ to predict the encoded edge map. Then we calculate the gradient of the similarity between the desired edges w.r.t. the input $\nabla_{z_t}\mathcal{L}$, and use it as a guidance for the denoising process that pushes the synthesized image to have edges close to the target edge map.}
    \label{fig:guidance_inference}
\end{figure}

\subsection{Sketch-Guided Text-to-Image Synthesis}
Given a sketch image $e$ and a caption $c$, our goal is to generate a corresponding highly detailed image that follows the sketch outline. Figure~\ref{fig:guidance_inference} illustrates the proposed latent features-based guidance described in detail below.

We start with a latent image representation $z_T$ sampled from a uniform Gaussian. Normally, the DDPM synthesis consists of $T$ consecutive denoising steps $z_t \to z_{t-1}$ which constitute the reverse diffusion process, with $z_0$ being an encoded output image. The reverse diffusion process, on each of the denoising steps $t = T, \dots, 1$, evaluates a density score gradient estimation $\varepsilon(z_t, t, c)$, and based on it, depending on a sampler algorithm, computes the next sample $z_{t - 1}$. Notably, the score gradient computation consists of the forward pass of the main denoising U-net model. Thus, once the quantity $\varepsilon(z_t, t, c)$ is computed, we may also collect the intermediate activations $l_1(z_t | t, c), \dots, l_n(z_t | t, c)$. 

Similarly to the training step, we concatenate these activations to a per-pixel spatial tensor $\mathbf{F}(z_t | c, t)$. Then using the pretrained per-pixel latent edge predictor $P$, to evaluate the step-$t$ latent edges prediction $\tilde{E}$ with $\tilde{E}_{i, j} = P(\mathbf{F}_{i, j})$. We can then calculate the similarity between the current prediction 
\begin{equation}
\mathcal{L}(\tilde{E},E(e)) = \|\tilde{E}-E(e)\|^2.
\end{equation}
Similarly to the external classifier gradient guidance in \cite{dhariwal2021diffusion}, we evaluate the anti-gradient $-\nabla_{z_t}\mathcal{L}$ to bring the edges-guidance to the diffusion process. Intuitively, this antigradient pushes an intermediate sample $z_t$ to have edges closer to the target. Now we replace the next-step sample prediction $z_{t - 1}$ with $\Tilde{z}_{t-1} = z_{t-1} - \alpha \cdot \nabla_{z_t}\mathcal{L}$, where $\alpha$ controls the edges guidance strength. In practice, the impact of this gradient depends on its relative magnitude to the original model step, hence, we normalize it with
\begin{equation}
\alpha = \frac{\|z_t - z_{t - 1}\|_2}{\|\nabla_{z_t}\mathcal{L}\|_2} \cdot \beta
\end{equation}
with $\beta$ being a constant throughout the synthesis process. Normally $\beta$ takes values of order $O(1)$. Once being synthesized with the guidance from the objective $\mathcal{L}$, the model produces a natural image aligned with the desired sketch.

In practice, as the final steps of the reverse denoising process commonly do not affect the geometric layout of the final generated image, we perform the edge guidance only for the steps $t = T, ..., S > 1$, where commonly $S = 0.5 T$. We further discuss the choice of the edge guidance stop step $S$ in the following section.
\section{Experiments}
\label{sec:exp}

\begin{figure*}
    \centering
  \includegraphics[width=0.999\textwidth]{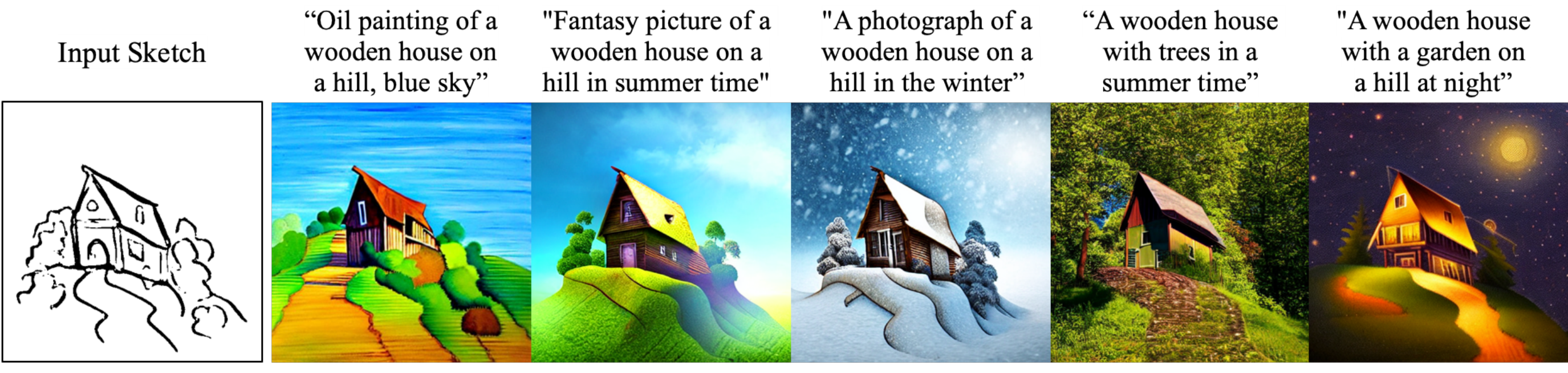} \\
  (a) \\
  \vspace{12pt}
  \includegraphics[width=0.999\textwidth]{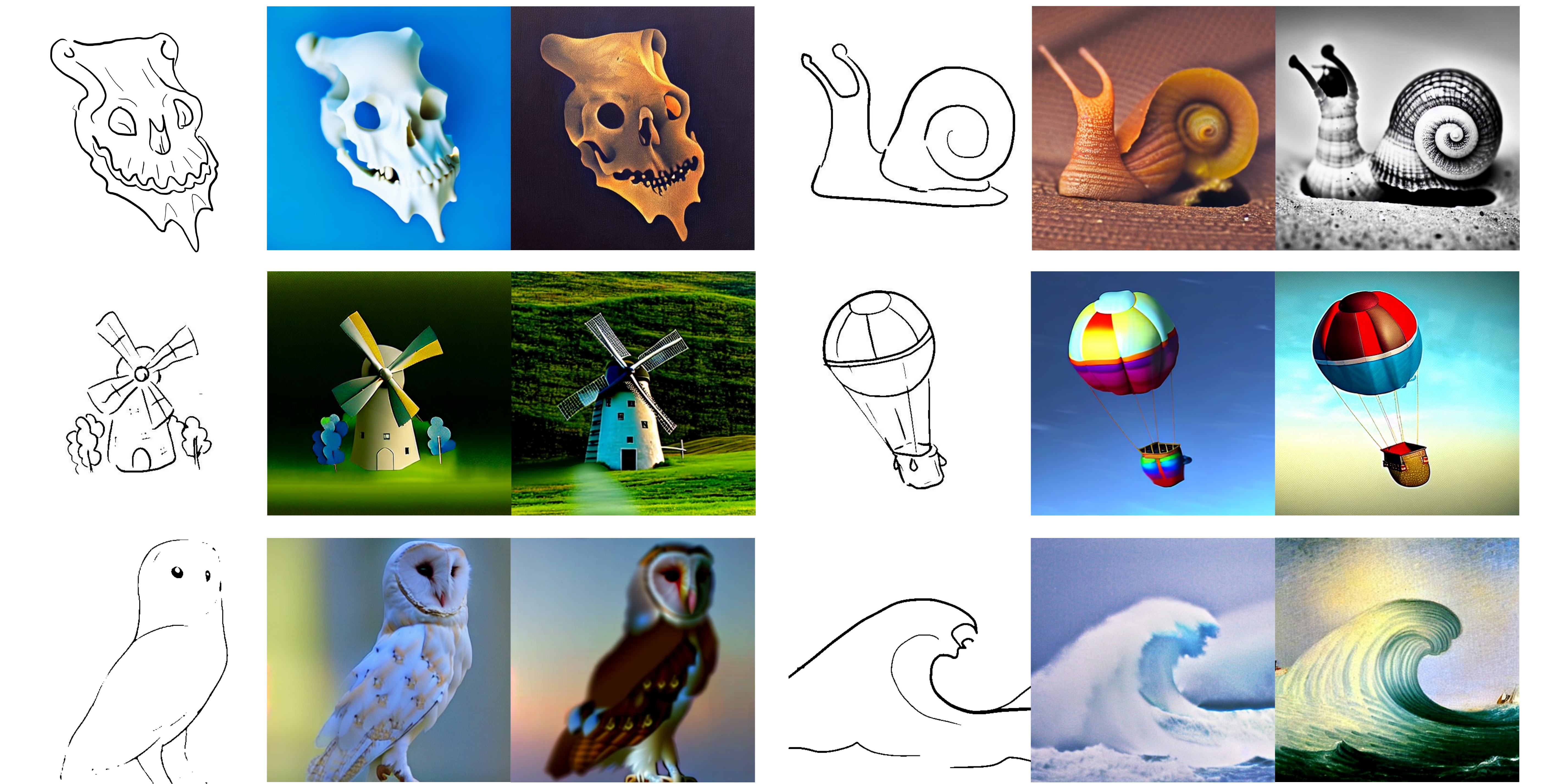} \\
  (b) \\
  \captionof{figure}{{\bf Sketch-Guided Text-to-Image Results.} (a) Samples of our method applied to input sketches with different prompts (depicted above the images). (b) More sketch-to-image results applied with different seeds and captions. It can be seen that our approach successfully handles objects from different domains. The corresponding captions as well as more results can be found in the supplementary material.}
  \label{fig:gallery}
\end{figure*}

In this section, we discuss the implementation details of our approach, show sketch-guided text-to-image synthesis results, conduct experiments and ablation studies to analyze the performance of various components in our framework, and present comparisons to state-of-the-art image translation techniques. Figure~\ref{fig:gallery} shows a gallery of results which
demonstrate the ability of our framework to convert sketches to images with an input text-prompt. 

\subsection{Implementation details}
In all of our experiments, we use Imagenet  \cite{deng2009imagenet} samples with their class names as captions (e.g., ``shoes"). The corresponding edge maps were generated with the edge prediction model of \cite{su2021pixel} and then thresholded with 0.5.
The latent edge predictor consists of 4 fully-connected layers with ReLU activations, batch normalization, and hidden dimensions $512$, $256$, $128$, $64$, and output dimension $4$. The denoising model's features are taken from 9 different layers across the network: input block - layers 2, 4, 8, middle block - layers 0, 1, 2, output block - layers 2, 4, 8. The training is performed for 3000 steps with Adam optimizer and batch size 16 which takes less than an hour on a single A100 GPU.

For inference, we found a set of reliable parameters for the edge guidance scale $\beta = 1.6$, guidance stop step $S = 0.5T$, and prompt-conditioning equal to $8$ (classifier-free guidance scale in DDPM), though these parameters can be modified based on the user requirement, to balance between edge fidelity and realism (see Section~\ref{sec:parameters} for more details).

\subsection{Comparisons}
We compare our method to three types of baseline approaches: SDEdit~\cite{meng2021sdedit}, pix2pix~\cite{isola2017image}, and PITI~\cite{PITI22}, each of which can be used for sketch-to-image synthesis.

\begin{figure}
    \centering
    \includegraphics[width=\columnwidth]{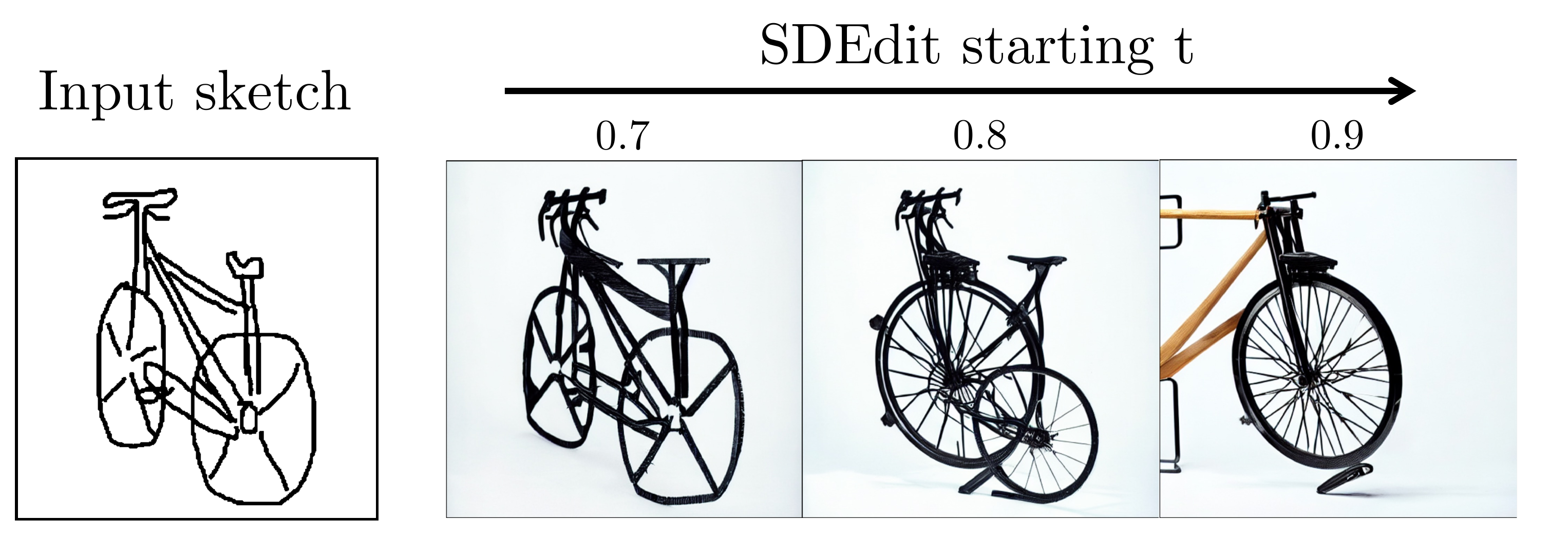}
    \caption{{\bf Applying SDEdit~\cite{meng2021sdedit} for Sketch-to-Image Translation.} For low starting-$t$ values, the system struggles to add colors and texture to the model, while for high starting-$t$ values the fidelity to the input sketch significantly decreases. Text-prompt used to condition the model: ``A photograph of a bike made of wood". }
    \label{fig:SDEdit}
\end{figure}

A possible attempt to solve the sketch-to-image task with a pretrained text-to-image diffusion model, is by adding noise to the input sketch, for $t$ steps in the forward diffusion process, then denoise it in a reverse process with a text prompt, as suggested in  SDEdit~\cite{meng2021sdedit}. This process enables to implicitly guide the model with a spatial map. However, as can be seen in Figure~\ref{fig:SDEdit}, the model expects that the guiding image lays in the RGB domain, hence, resulting in unnatural, black and white images that follow the input sketch (text-prompt condition used: ``A photograph of a bike made of wood"). For low values of $t$, the system struggles to add texture to the model, and when $t$ is increased, the fidelity to the input sketch significantly decreases.

\begin{figure}
    \centering
    \vspace{-10pt}
    \includegraphics[width=\columnwidth]{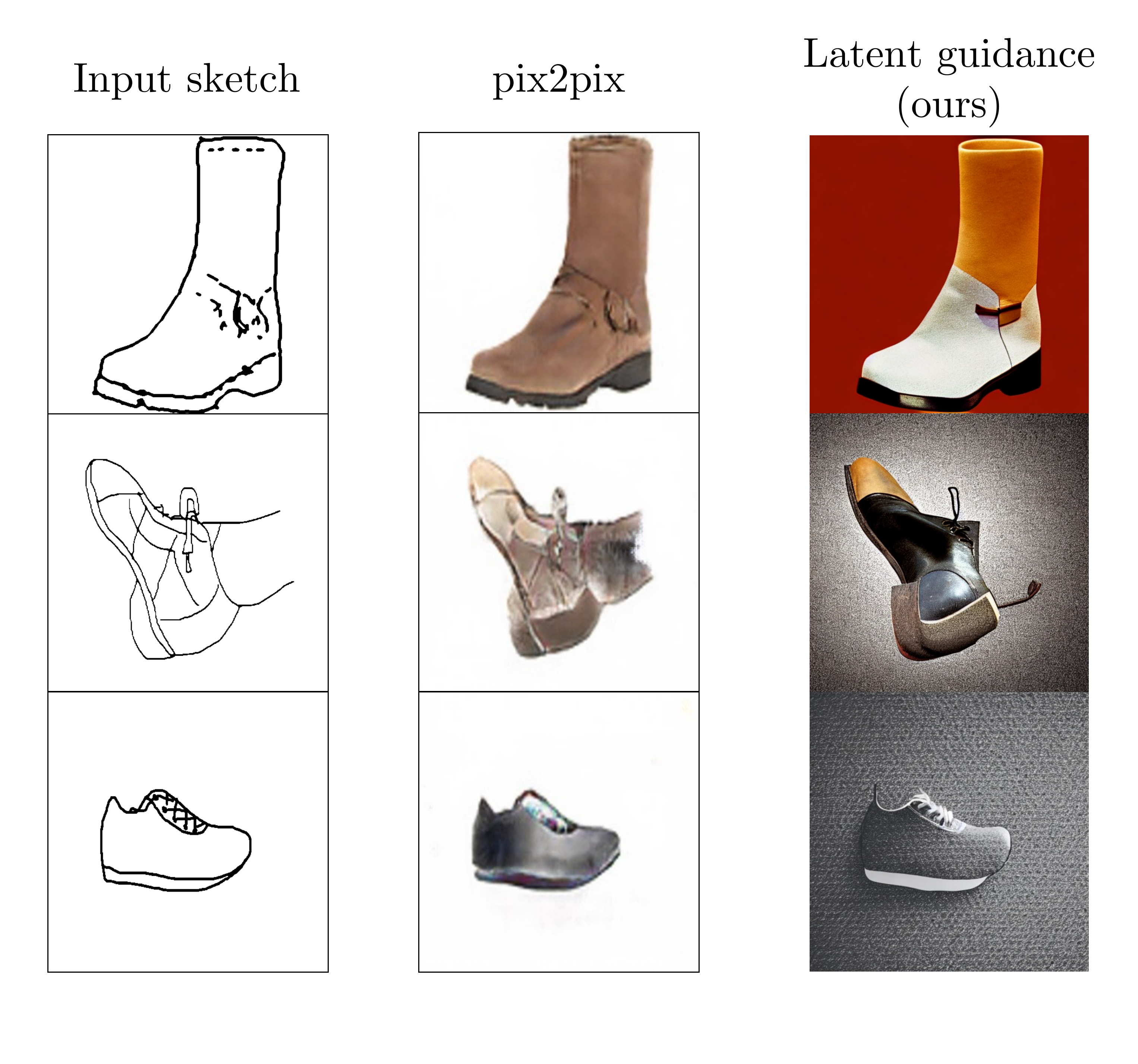}
\vspace{-20pt}
    \caption{{\bf Comparison to pix2pix~\cite{isola2017image}.}  pix2pix works well only on sketches that lay within the domain of the training data, but fails on out-of-domain hand-drawn sketches. 
}
\vspace{-12pt}
    \label{fig:sbs_shoes}
\end{figure}

We next compare our approach to pix2pix~\cite{isola2017image}, and PITI~\cite{PITI22}. Pix2pix is a self-supervised method that is trained on pairs of images (in this case real images and their corresponding sketches) using a reconstruction loss that is enhanced by the adversarial loss that is applied to pairs. Figure~\ref{fig:sbs_shoes} demonstrates that this approach works well on sketches that lay within the domain of the training data, for example, realistic shoes sketch, while failing on out-of-domain hand-drawn sketches. 

\begin{figure*}
  \centering
  \includegraphics[width=0.999\textwidth]{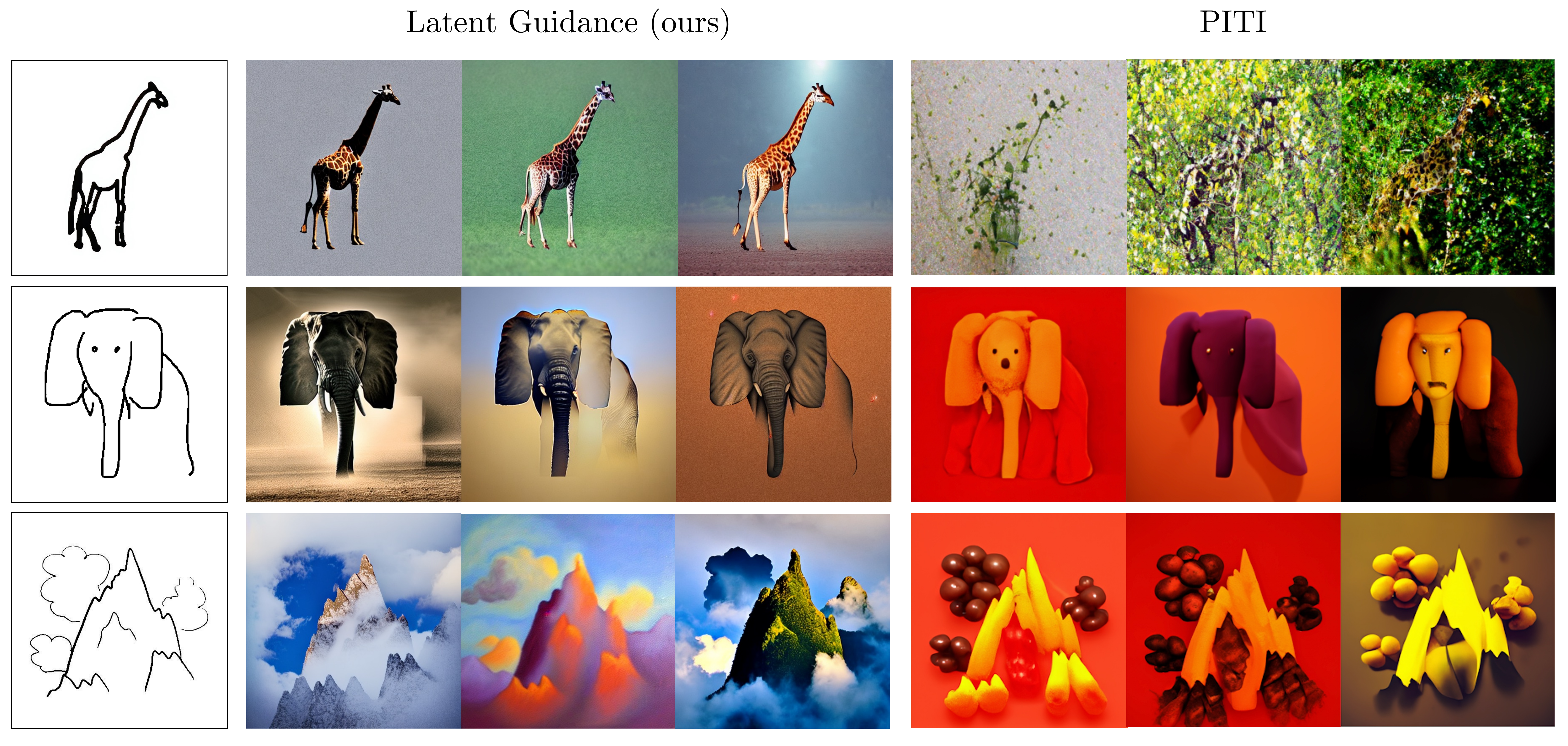}
  \caption{{\bf Comparison to the state-of-the-art approach of PITI~\cite{PITI22}}. While PITI performs well on realistic sketch samples, it struggles to create realistic outputs on free-hand sketches that are out of their training data domain. In addition, notably, our method provides significantly more color and style variability compared to PITI.}
  \label{fig:sbs_elephant}
  \vspace{-10pt}
\end{figure*}

PITI~\cite{PITI22} trains a dedicated encoder to map the guiding image into the
latent space of the pretrained unconditional diffusion model. 
Figure~\ref{fig:sbs_elephant} shows that while PITI performs well on realistic sketch samples, it struggles to create realistic outputs on free-hand sketches that are out of their training data domain. In addition, notably, our method provides significantly more color and style variability compared to their approach.

\subsection{Ablations and Parameter Tuning}
\label{sec:parameters}
\begin{figure}
    \centering
    \vspace{-10pt}
    \includegraphics[width=\columnwidth]{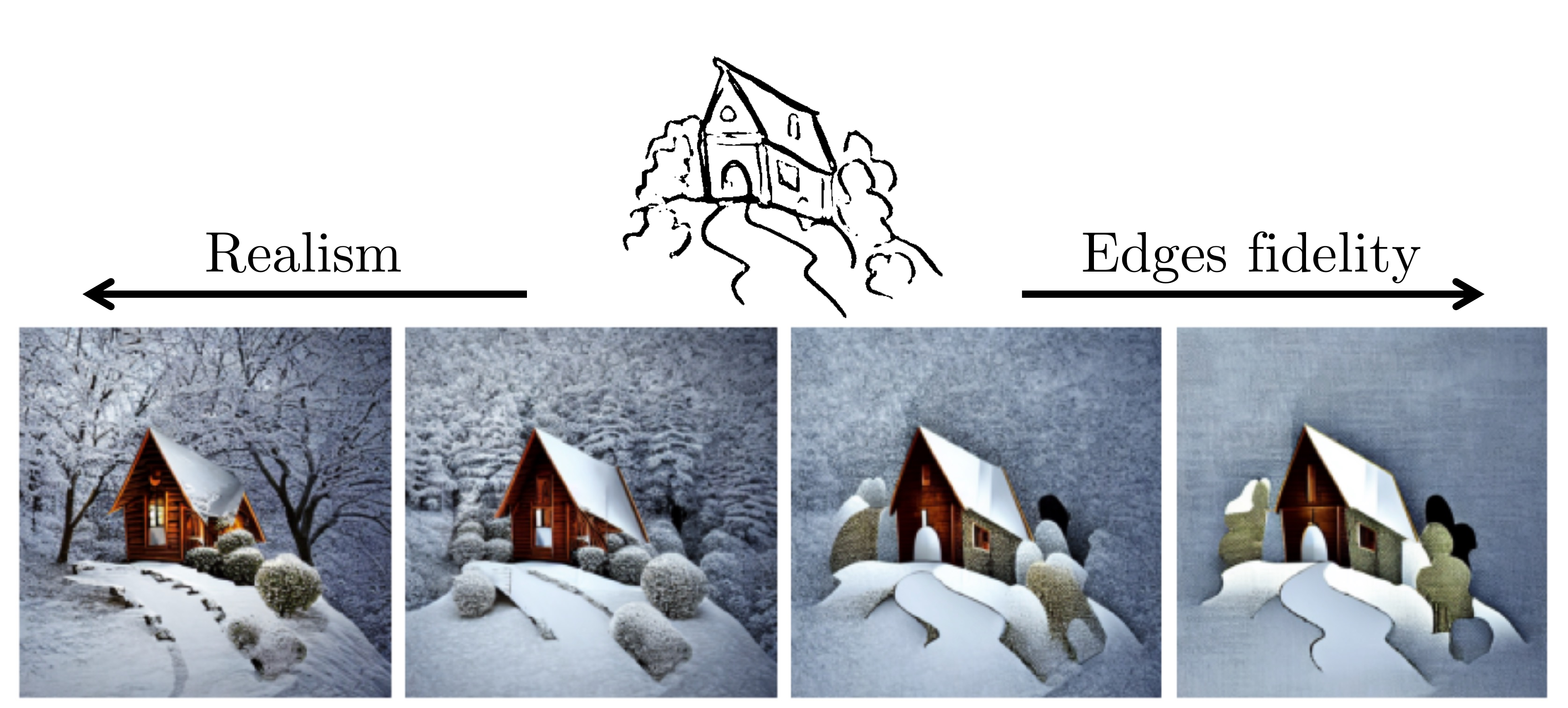}
    \caption{{\bf Realism vs Edge-Fidelity.} Our method demonstrates a trade-off between the realism level of a generated image and its alignment with the edges of the target sketch. The trade-off can be controlled by the edge-guidance scale $\beta$.}
    \label{fig:real_edge_tradeoff}
    \vspace{-10pt}
\end{figure}

We next discuss the different parameters in our system and their effect on edge fidelity.
First, our method demonstrates a trade-off between the realism level of a generated image and its alignment with the edges of the target sketch. The trade-off, which can be controlled by the edge-guidance scale $\beta$, is depicted in Figure~\ref{fig:real_edge_tradeoff}. It can be seen that for small values of $\beta$, we get a more realistic image with details and textures that cover regions in the entire image, while the larger value of $\beta$ favors edge alignment but generates less realistic, piece-wise smooth, results.

We also quantitatively measured the edge-fidelity (Mean Squared Error between the target edge map and the edges of the synthesized image), as a function of the guidance stop step $S$ and depicted the result in Figure~\ref{fig:target_edge_fidelity}. As expected, the quality of edge reconstruction is improved for larger values of $S$. 
However, since high edge fidelity comes at the expense of realism, we want to find a sweet spot that will enable us to balance these two factors. For that, we conducted an experiment that measures the reconstruction error of our Latent Edge Predictor for different values of $t$. Figure~\ref{fig:edge_error} depicts the loss in Equation~\ref{eq:loss_main} as a function of $t$ . Notably, starting from $t \approx 0.5T$, the error stabilizes, indicating that for  $t < 0.5T$, the model does not receive new information on the edges. Hence, we use this stabilization point as the guidance stop $S$ to mitigate the trade-off, which is highly aligned with the segmentation errors as a function of $t$ that were reported in \cite{baranchuk2021label}, that uses latent features of DDPM for the few-shots semantic image segmentation task. 

\begin{figure}
    \centering
    \includegraphics[width=\columnwidth]{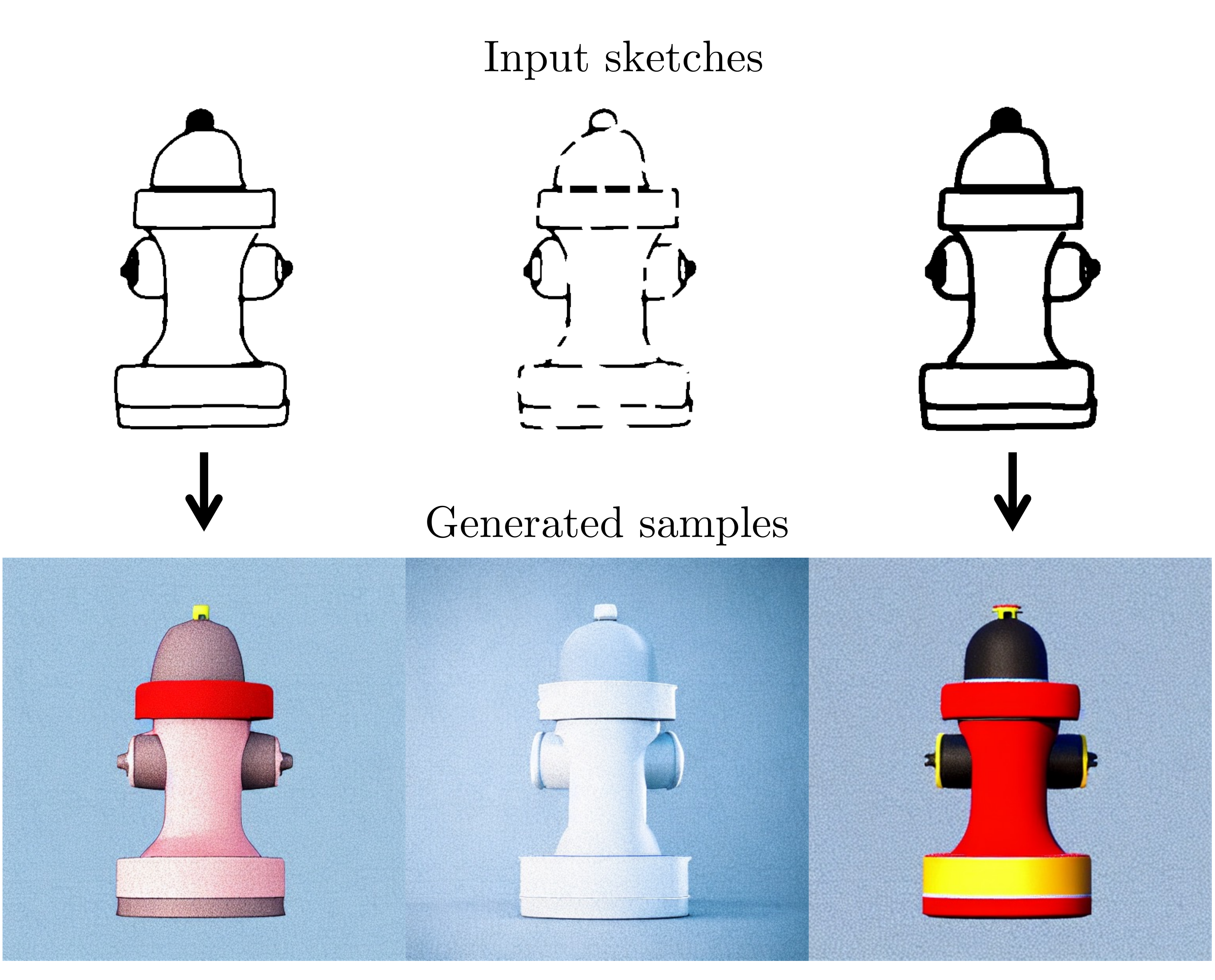}
    \caption{{\bf Stroke Ablation.} Samples generated with the same sketch geometry but different stroke styles, yield the same shape but with variation in colors and textures.
    It demonstrates the insensitivity of our approach to the strokes styles.}
    \label{fig:stroke_ablation}
    \vspace{-10pt}
\end{figure}

\begin{figure}
    \vspace{-10pt}
    \centering
    \includegraphics[width=\columnwidth]{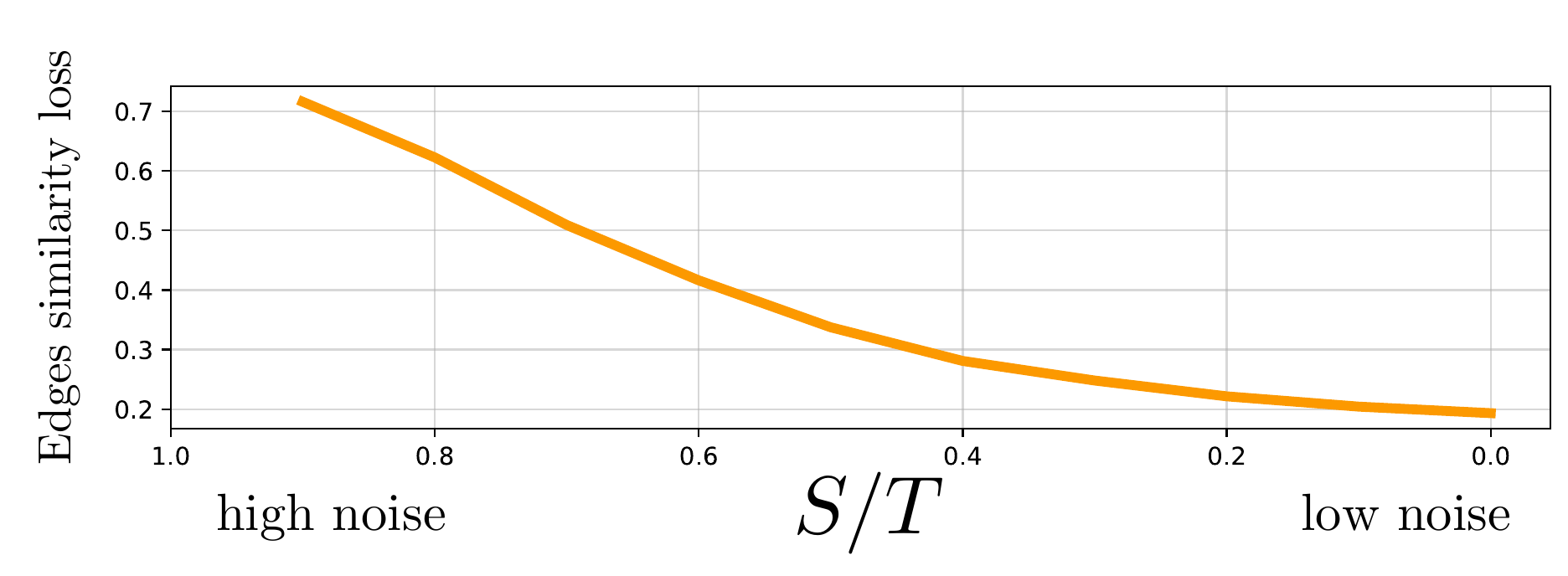}
    \caption{{\bf Edge-fidelity as a function of guidance stop step $S$. } The edge fidelity of the output image (to the target edge map) as a function of the normalized guidance stopping step $S/T$.}
    \label{fig:target_edge_fidelity}
    \vspace{-5pt}
\end{figure}

\begin{figure}
    \vspace{-10pt}
    \centering
    \includegraphics[width=\columnwidth]{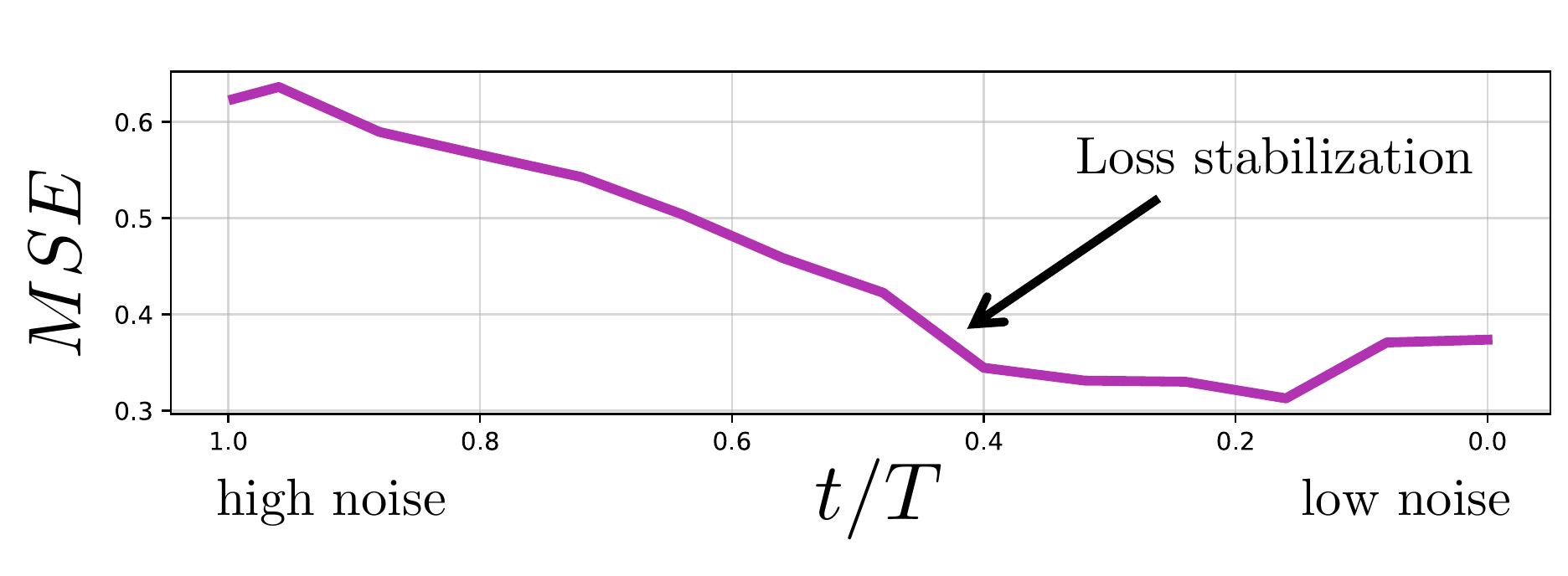}
    \caption{{\bf   Reconstruction Error of the Latent Edge Predictor.} Reconstruction Error (MSE) as a function on the normalized noise level $t/T$ of the diffusion process.}
    \label{fig:edge_error}
\end{figure}

Since our latent edge predictor works in a local, per-pixel, manner, we also demonstrate its insensitivity to the stroke style. We generated samples produced with the same sketch geometry but with different stroke styles.  Figure~\ref{fig:stroke_ablation} shows that such a setting yields the same shape, but with variation in colors and textures. This observation explains that the stroke style affects the inner synthesis process only, which accumulates into varying colors of the output image.

\section{Applications}
We demonstrated our spatially-guided text-to-image synthesis approach on the sketch-to-image application, however, our approach is generic and can be applied to different image-to-image translation tasks. In this section, we show how  can we use saliency maps as a guiding map for text-to-image models, and how it can be used for natural enhancement of image regions, as well as background inpainting~\cite{kim2022zoom}. More applications and examples can be found in the supplementary material.

\subsection{Saliency Guidance}
\begin{figure}
    \centering
    \vspace{-10pt}
    \includegraphics[width=\columnwidth]{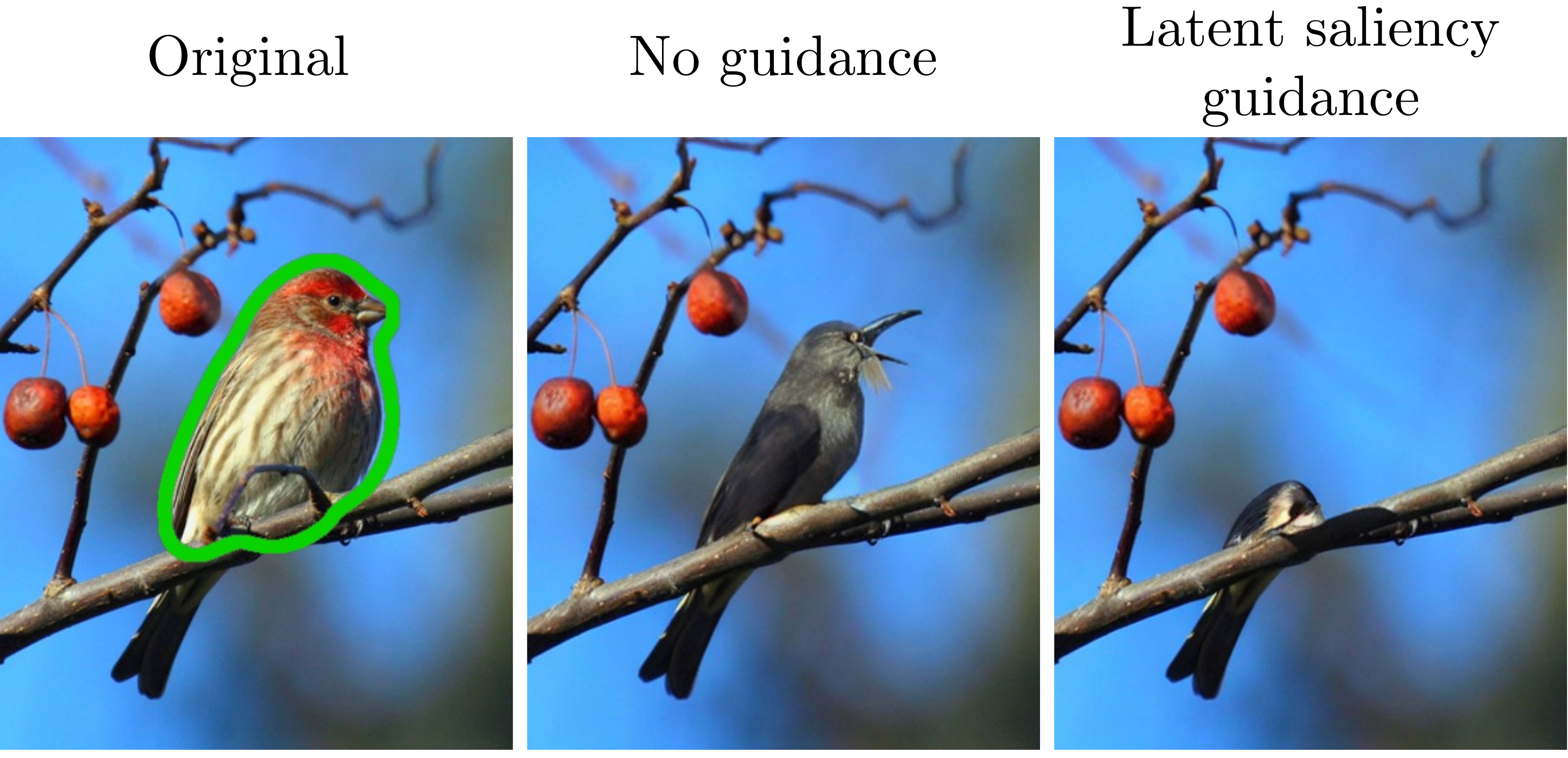}
    \vspace{-5pt}
    \caption{{\bf Background Inpainting}. Our approach can use a saliency map to guide the synthesis process of diffusion models when applied to image inpainting. Given an image (leftmost) and a mask ((marked in green border), when inpainting is being solved with the default settings of the model (no guidance) the region may be filled with foreground subjects based on the semantic hints provided in the non-masked regions, like the tail of the bird (middle). In contrast, when the guiding map has low saliency values in the masked region the completed region contains background values  (rightmost).}
    \label{fig:saliency_down}
    \vspace{-15pt}
\end{figure}

\begin{figure}
    \centering
    \includegraphics[width=\columnwidth]{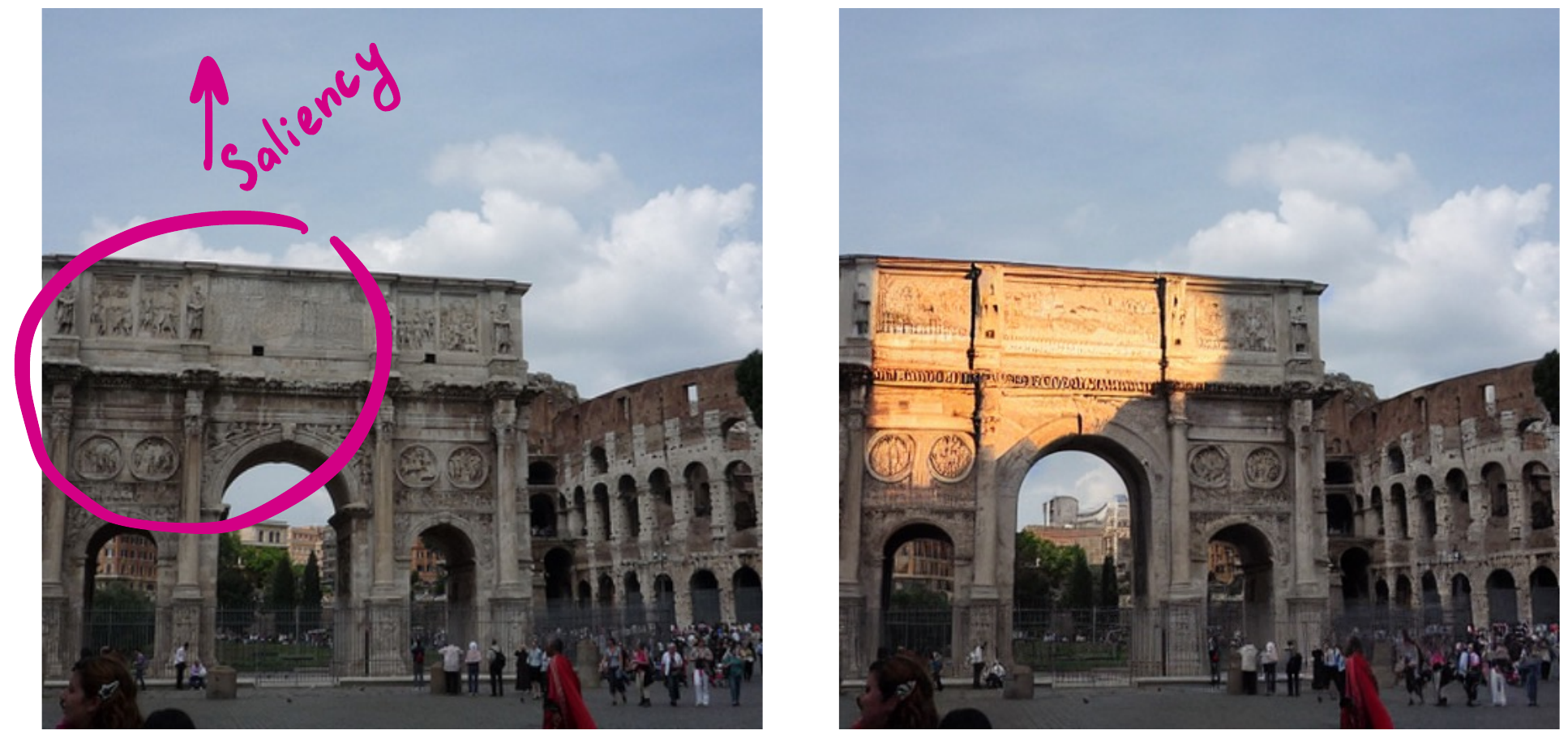}
    \caption{{\bf Saliency-Guided Text-to-Image Synthesis}. Given an image (left) and a saliency map with high values in a specific region (marked in pink sketch), our model modifies the values within the image to highly be illuminated by the sun due to the requirement for high saliency values in this region.}
    \label{fig:saliency_up}
\end{figure}

Saliency prediction models can be used to detect the most attention grabbing regions within an image. Recent works have shown that saliency can be also used as a guiding component for image editing, to reduce distraction in images~\cite{aberman2022deep}. We next show that saliency maps can also guide text-to-image diffusion models such that the saliency in specific areas of the generated image is high or low. 
In this case, we train our latent guidance predictor to directly predict the original (not encoded) downsampled saliency maps from our noisy latent features, and we optimize the model with the binary cross-entropy loss instead of mean squared error. To supervise the MLP we use the saliency model from~\cite{U2Net}. We then run the inpainting model with out-of-mask conditioning, and with an empty prompt. Figure~\ref{fig:saliency_up}. demonstrates how this technique can be used for background inpainting. For a given image of a bird and a mask that covers the bird's body without the tail, it can be seen that the model fills the hole with a new bird due to the semantic hint that the tail provides to it. In contrast, when the model is guided by a saliency map with low values in the mask region, it simply removes the bird body and fills it in with background inpainting as expected. In addition, we can guide the model to generate high saliency values within a region. Figure~\ref{fig:saliency_down} shows how the marked region is highly illuminated by the sun due to the requirement for high saliency values in this region. Notably, it is sufficient to apply the latent guidance predictor for only the first $20\%$ of the steps.

\section{Conclusions}

We presented a technique to guide a pre-trained text-to-image model diffusion model with a spatial map. We have focused on sketch-guidance, and showed that the technique can handle well out-of-domain sketches, which may have a large variety of styles completely different than those seen in the training time. The gist of the technique is the per-pixel training of a lightweight MLP component that is trained on rather small training data. The per-pixel training acts more like a differential edge-detector and unlike common per-image training, it is not bound to a particular global sketching style.

Our technique piggybacks on a pertained text-to-image model diffusion model and thus offers a strong multi-modal sketch-guidance technique to users. In a sense, the technique accepts a rich variety of sketching styles and at the same time provides a rich variety of outputs, where the user has intuitive control over the input, and semantic control over the output.

\begin{figure}
    \centering
    \includegraphics[width=\columnwidth]{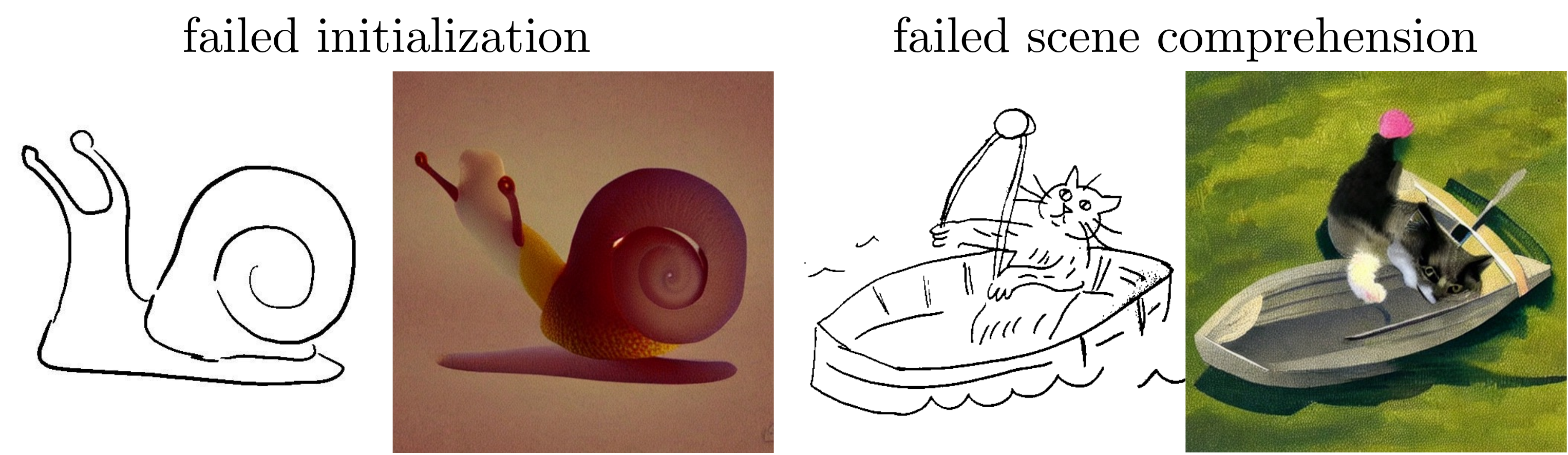}
    \caption{{\bf Failure cases.} The quality of the results may drop for different initialization, and on complex scenes with mixed and ambiguous semantics.}
    \label{fig:fails}
    \vspace{-8pt}
\end{figure}

Still, our presented technique is only a step toward gaining more control over the output of generative text-image models. The technique has its limitations. Currently, the technique is vulnerable to the local style of the strokes. The technique still struggles with complex and cluttered sketches as it treats all of the strokes equally without prioritizing them according to their saliency or semantics. Also, since the text-image diffusion model is stochastic, there might be conflicts between the random seed and the input sketch, which may lead to a generation of an output that does not agree well with the sketch. Figure~\ref{fig:fails} shows representative examples where the model fails to provide satisfying results. The quality of the results may drop for different initialization, and complex scenes with mixed and ambiguous semantics.

In the future, we would like to advance and improve the technique by adding a sketch inversion step to yield a stronger seed to the diffusion process, to better push the output toward the outline of the input sketch. Another direction is to quickly learn a personalized style using just a few shots. With a quick training session, the latent sketch predictor can accommodate the artist's stroke style.

\section{Acknowledgements}
We thank Chu Qinghao, Yael Vinker, Yael Pritch, Dani Valevski and David Salesin for their valuable inputs that helped improve this work.

{\small
\bibliographystyle{ieee_fullname}
\bibliography{egbib}
}

\newpage

\appendix
{\Large{\centering\textbf{Supplementary material}}}
\section{Societal Impact}

This work provides a powerful tool to convert simple sketches to detailed, highly realistic images with full control over style and content. As reported by professional artists we interviewed, while a simple sketch drawing takes just a few minutes, a detailed colored picture based on it normally takes way more time and counts in hours. Thus, this work can potentially significantly speed up the process of artistic creation, enabling the democratization of creativity. This work is inspired by the idea of \textit{not replacing an artist, but giving an artist a tool} where AI takes all the technical parts of the creativity process while leaving the imagination and inspiration to a human.
As the back side of the proposed method, this also gives a tool for deep-fake and misleading material creation, this once more stands the challenge before the community to create the safety mechanism to prevent the generative models to be used in with controversial intentions.

\section{Ablation studies}

We start by performing a test to highlight the "out-of-domain" virtue of the proposed technique.
Our Latent Edge Predictor (LEP) appears to perform well out of its training domain samples due to its per-pixel training nature. We examine it with an extremely tiny training set. We train the edges predictor on the subset of Imagenet validation set, consisting of the dogs' classes only. The qualitative evaluation and comparison of a model trained with this single-domain protocol is depicted in Figure \ref{fig:limited_data}. Notably, even in this minimal setup, it still performs reasonably well.

To highlight the importance of using the deep features of the diffusion network, rather than the intermediate states, we perform the following experiment.
We train an independent noise level-conditioned edge predictor that operates over intermediate states $z_t$, instead of the internal features of the network. Though this is a straightforward generalization of the classifier-guidance \cite{ho2021classifier} technique, we have not succeeded to train a plausible predictor, as it commonly collapses to the prediction of empty maps. We argue that this is due to the fact that edge prediction based on a noisy image is almost as complex as the original DDPM denoising, which requires a comprehensive image understanding. Thus, making this approach work might take a significant computational effort, while our proposed per-pixel training works out of the box for a rich variety of tasks. In addition, our approach takes only an hour to train and requires a limited amount of data.

\begin{figure}
    \centering
    \includegraphics[width=\columnwidth]{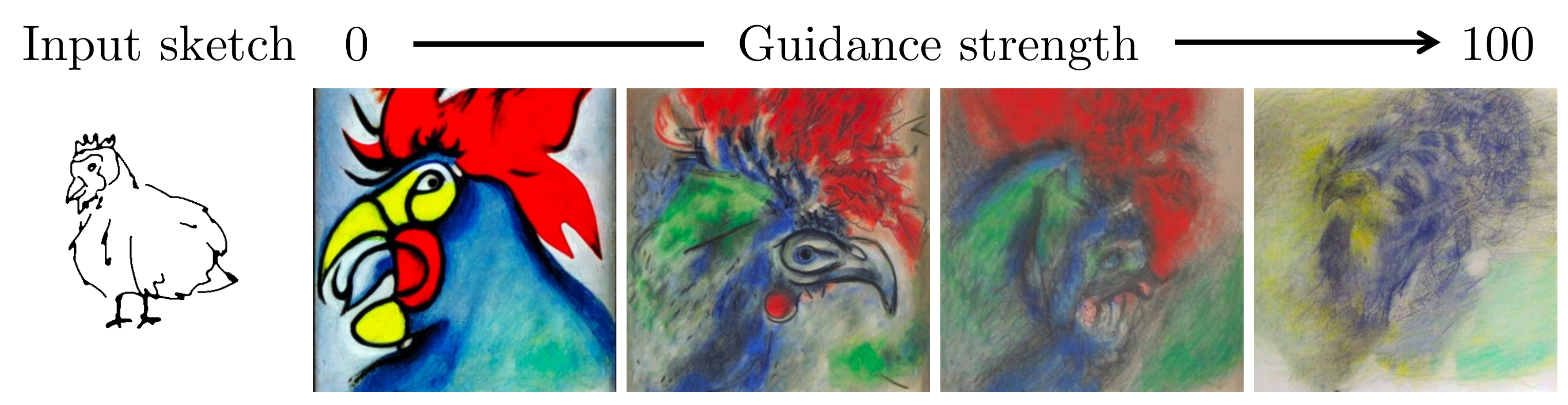}
    \caption{{\bf Guidance with a model that operates on the intermediate predictions of the denoised image.}  Generation with low guidance weight fails to produce an image that matches the sketch, while high guidance scale produces nearly adversarial images.}
    \label{fig:non_conditioned_MLP}
\end{figure}

We also tried to perform guidance over the intermediate $z_0$ predictions of the DDPM model -- a more intuitive input to the edge predictor which operates better on clean images. In each of the denoising steps $z_t \to z_{t-1}$, the model simultaneously predicts the end result $z_0 = z_0(t)$. Given this prediction, we compute the current edges prediction as $e(E^{-1}(z_0(t)))$ where $E^{-1}$ states for the VQVAE decoder, and $e(\cdot)$ is the edge prediction model we use for the edges labeling. Then we guide the denoising process with the gradients of the similarity of the predicted edges and target edges. Despite the fact that the edge predictor operates on a noise-free image, it still struggles to predict edges from those images that are directly estimated from fully noisy images, rather than passing through the entire diffusion process. Hence, the entire process of sketch guidance fails. Figure~\ref{fig:non_conditioned_MLP} demonstrates the guidance performed that way. It can be seen that samples with low guidance weight fails to produce an image that matches the sketch, while high guidance weight produces nearly adversarial images.

\section{Additional tests}

\paragraph{Leveraging the Multimodality}
Our approach enables to use inputs from two different modalities - a spatial sketch map and a text. Figure \ref{fig:cows} visualizes the effect of different prompts guidance. While an empty prompt commonly induces unsatisfying results, a minimal relevant prompt induces plausible generation. Commonly, a detailed description of a sketch subject induces more realistic and detailed generation (column 4: "...a cow on a snowy field..."). Once a prompt contains objects non-presented in the sketch, it may confuse the generation process (in column 5, "...a cow surrounded by trees.": there are no tree edges presented in the sketches. While the model succeeded in generating a proper environment for the second sketch, for the first sketch it makes the trees to be formed by the sketch subject shape). Adding an artistic prefix (column 6: "An oil painting...") always induces high alignment with the sketch as in that case the prompt guidance is less concerned about the generated image realism. The rightmost column shows conflicting text prompt ("a dog"). The output is indeed a dog that admits to the guiding sketch. This clearly highlights the competence of our edge-guiding technique.

\begin{figure}[h!]
    \centering
    \includegraphics[width=\columnwidth]{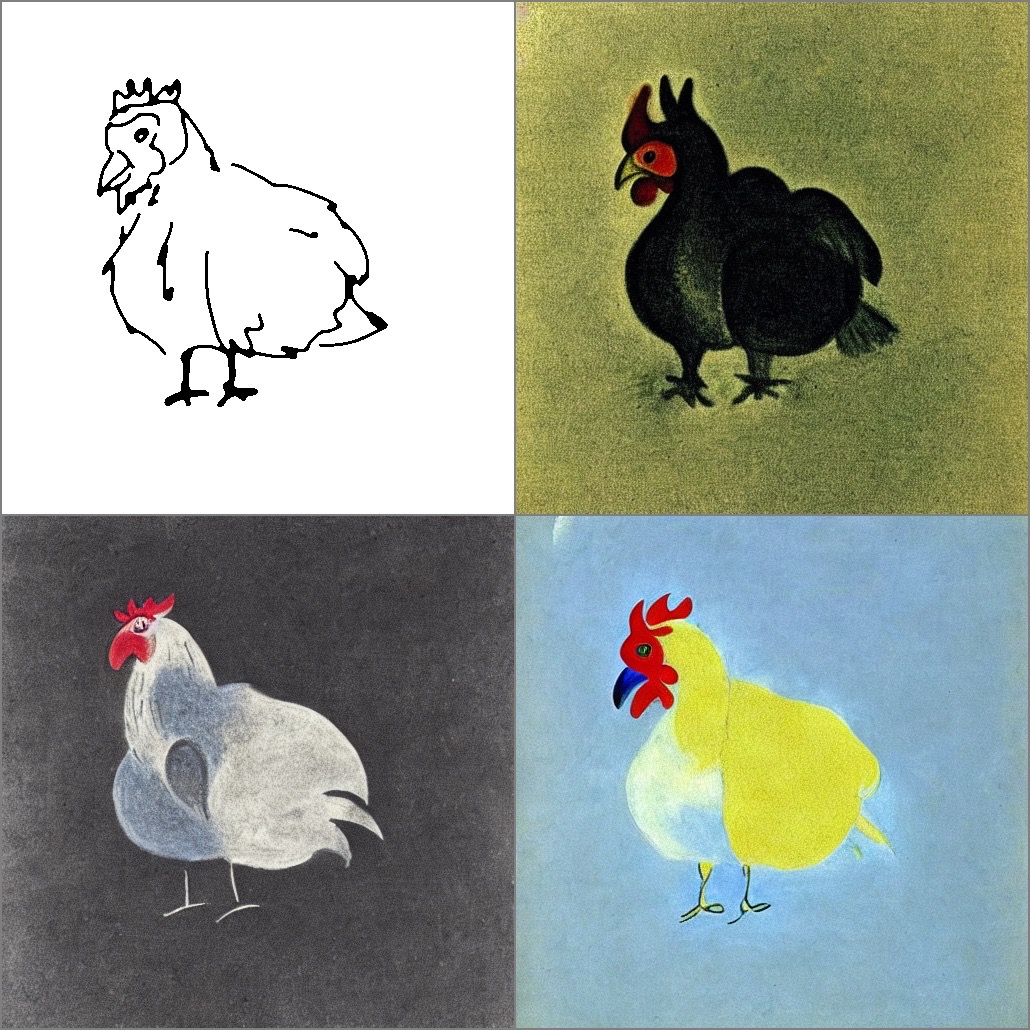}
    \vspace{1pt}

    \includegraphics[width=\columnwidth]{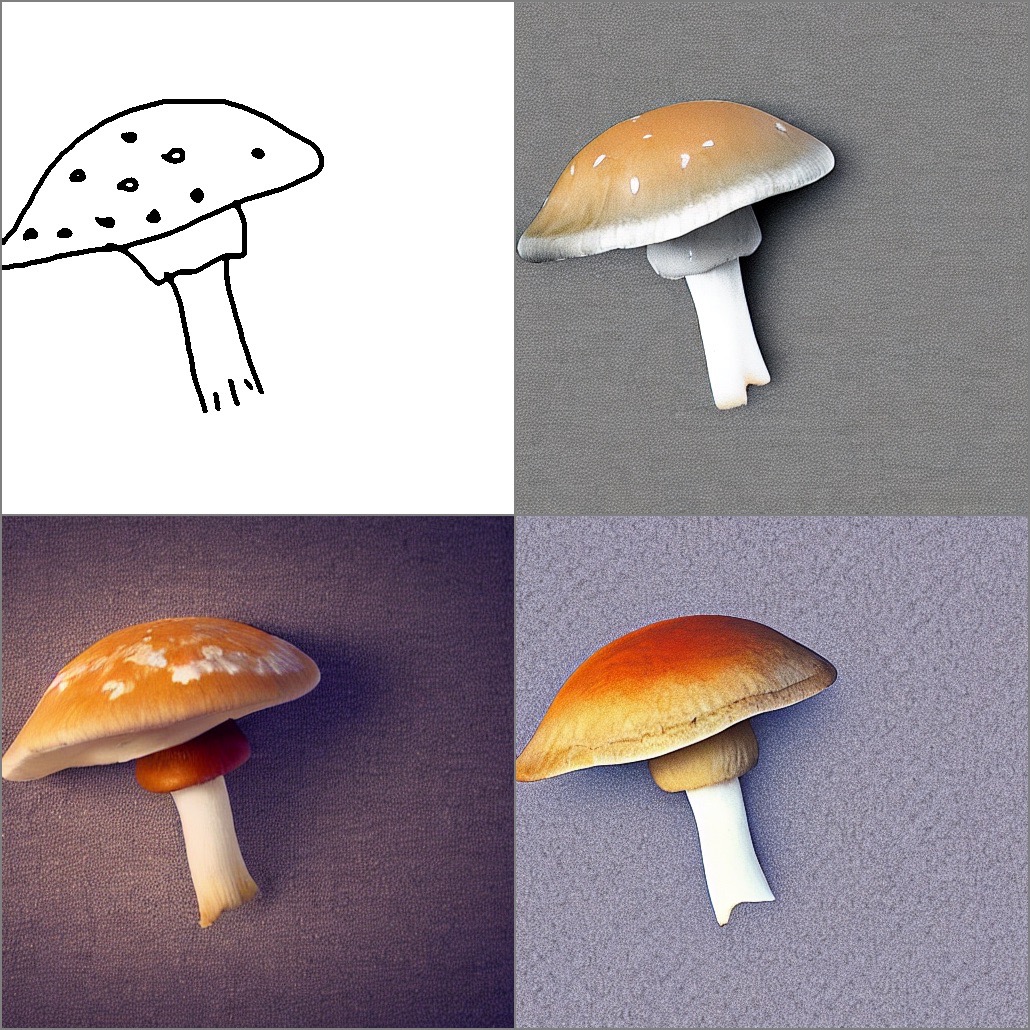}
    \caption{{\bf Sampling with different seeds.} A variety of different samples generated with the same sketch and a prompt, but different seeds. Top sketch prompt: "Marc Chagall drawing of a rooster.", bottom sketch prompt: "A photograph of a mushroom."}
    \label{fig:resampling}
\end{figure}

Note that due to stochastic sampling of DDPM, a single sketch and a prompt can generate a variety of different samples, as shown in Figure \ref{fig:resampling}.

\begin{figure}
    \centering
    \includegraphics[width=\columnwidth]{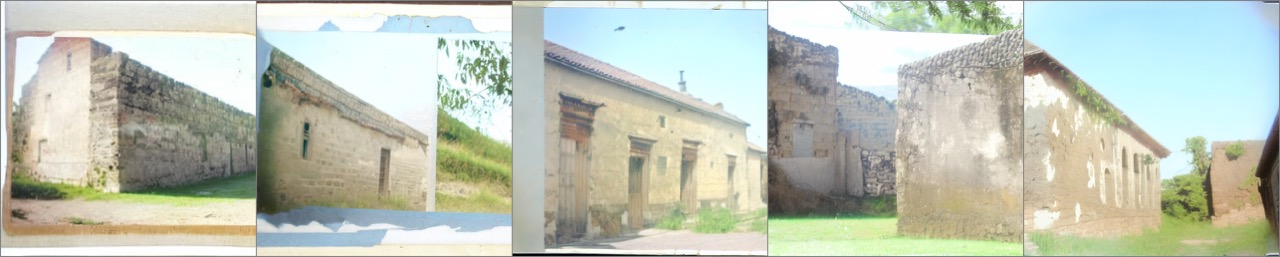}
    \includegraphics[width=\columnwidth]{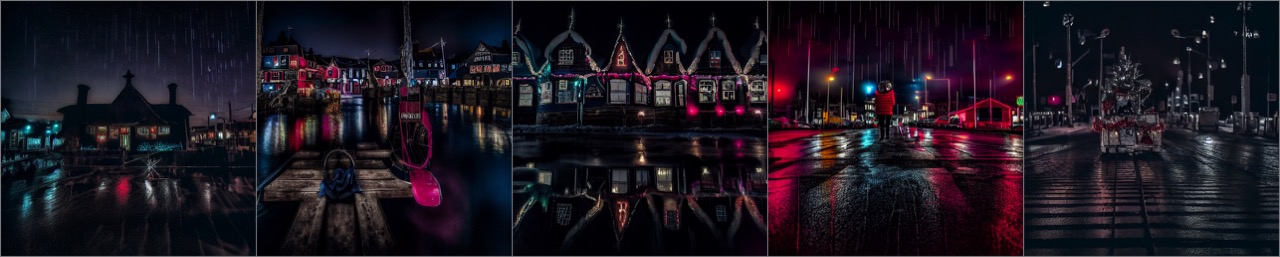}
    \includegraphics[width=\columnwidth]{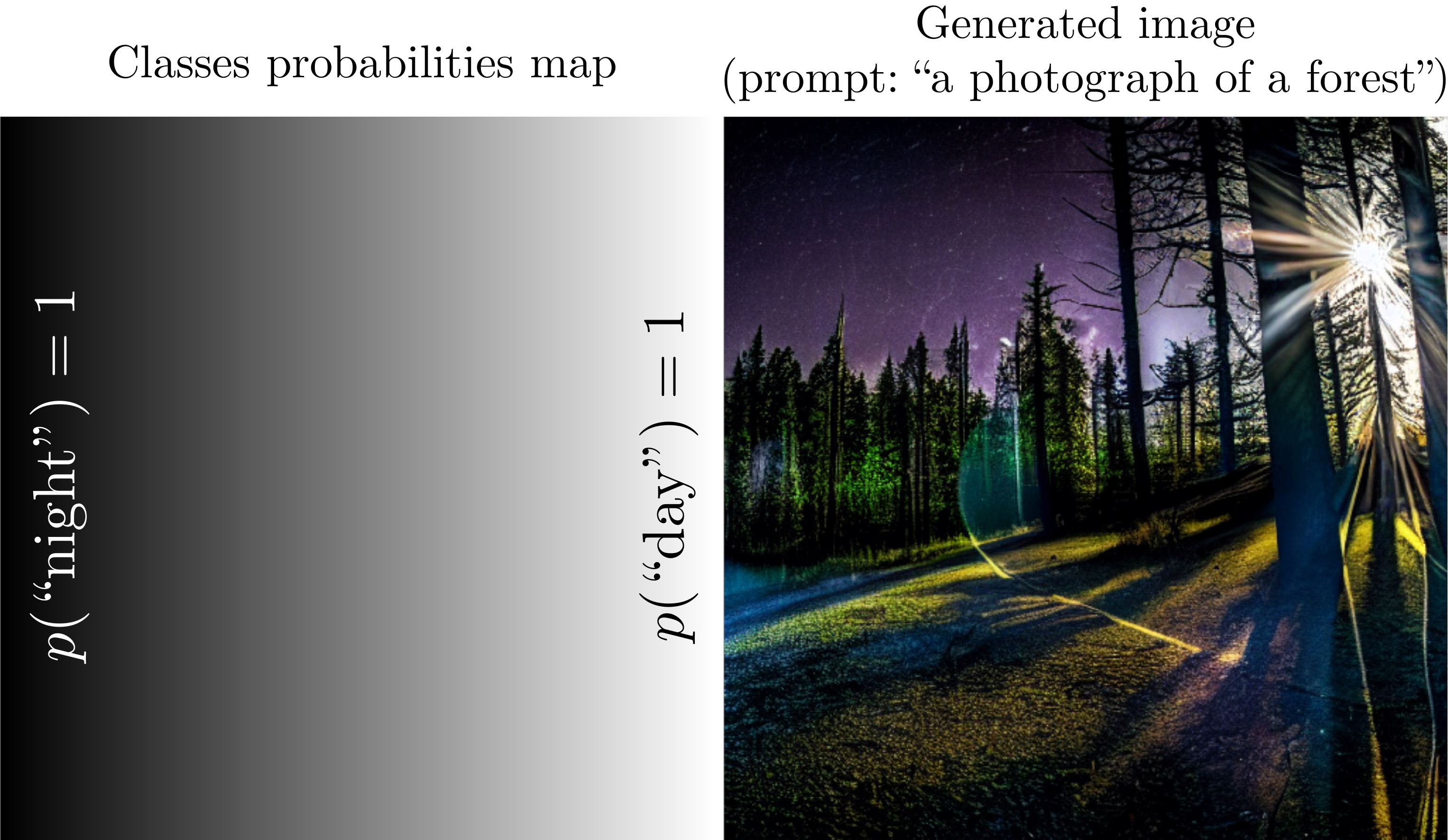}
    \caption{{\bf Guidance of a Spatial Label Map.} Top rows: images generated with the class guidance map being equal to "day" and "night" (first and second orw, respectively) with the prompt: "a photograph of an old city". Bottom: image generation with a spatially varying soft label map. While the right side of the image contains a bright sun, there are stars and a black sky on the left side.}
    \label{fig:clsf_map}
\end{figure}

\begin{figure*}
    \centering
  \includegraphics[width=0.999\textwidth]{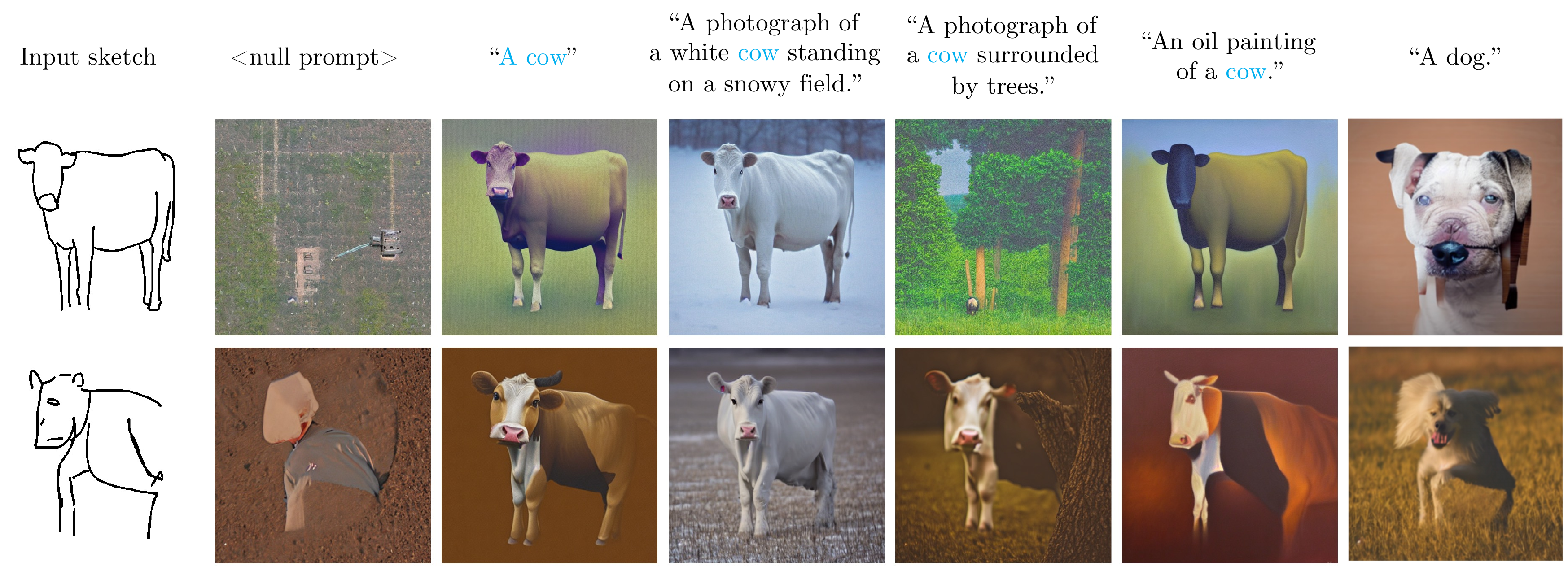}
    \caption{{\bf The effect of the textual conditioning on the sketch-guided generation.} The blue text highlights the subject presented in the sketch. The rightmost column shows a generation failure case where the prompt not matching the sketch.}
    \label{fig:cows}
\end{figure*}

\paragraph{Spatial Labels Guidance}
We next show another application of our generic approach and use a soft $[0,1]$ grayscale map to guide the diffusion process, where the maximal and minimal values of the map represent two classes - day and night. We train our per-pixel MLP to predict whether a pixel belongs to a day or night scene. Namely, based on the aggregated noised stacked noised features $\mathbf{F}(z_t | c, t)_{i, j}$, the MLP model $P$ predicts either the spatial location $(i, j)$ on the original encoded image $z$ represent a scene at the day time, or at the night time. We train $P$ with only 500 day and 500 night images, where all pixels from the one image (day or night) corresponded to the same class. As the data is limited, we perform optimization with 1000 steps only. The training scheme remains unchanged except for the loss where now we use the cross-entropy between $P(\mathbf{F}(z_t | c, t)_{i, j})$ and the ground-truth class at the location $(i, j)$. We also always use the null prompt for feature extraction. Similarly, throughout the generation, we guide with the cross-entropy loss. Now, when the guidance is performed with a constant 1 or 0 spatial labeling map, the produced images are either attributed to day or night (Figure~\ref{fig:clsf_map}, top). When the labeling map is formed by the interpolation between labels probabilities, the generated image also interpolates the scene between night and day (Figure~\ref{fig:clsf_map}, bottom).

The proposed method induces a computational overhead that is mostly induced by the backpropagation of the edge prediction loss from the inner features to the input image. Once the guidance is applied for the first half of all reverse diffusion steps, the relative sampling time overhead is approximately 80\%.

\section{Models details and data}
Dataset: Sketches presented in Figure 4, Figure 3 (third row), and Figure 14 are provided by the authors. The sketch presented in Figure 6 (first row) is taken from edge2shoes dataset \cite{isola2017image}, and the rest of the samples are taken from Sketchy dataset \cite{sangkloy2016sketchy}. All synthetic quantitative results are based on Sketchy dataset, and the real numbers are based on Imagenet with class names used as prompts.

Prompts used in Figure~4~(b): "A skull of a monster.", "A macro photograph of a snail.", "A photograph of a windmill.", "A hot air balloon.", "A photograph of a barn owl.", "A photograph of a big wave." (left in the last row), "William Turner's picture of a big wave.". A prompt used in Figure~6: "A shoe.". Prompts used in Figure~7: "A photograph of a giraffe.", "A photograph of an elephant.", "A mountain in clouds.", "An oil painting of a mountain in clouds.", "A photograph of a green mountain in clouds.". Figure~8: "A photograph of a wooden house on a hill in the winter.". Figure~9: "A hydrant.".  

In all the experiments except inpainting, we use the Stable Diffusion checkpoint \href{https://huggingface.co/CompVis/stable-diffusion-v-1-4-original}{stable-diffusion-v-1-4-original}. As for inpainting, we use the checkpoint \href{https://huggingface.co/runwayml/stable-diffusion-inpainting}{stable-diffusion-inpainting}. We always sample in the non-deterministic mode with 250 reverse diffusion steps.

\begin{figure*}
    \centering
  \includegraphics[width=0.92\textwidth]{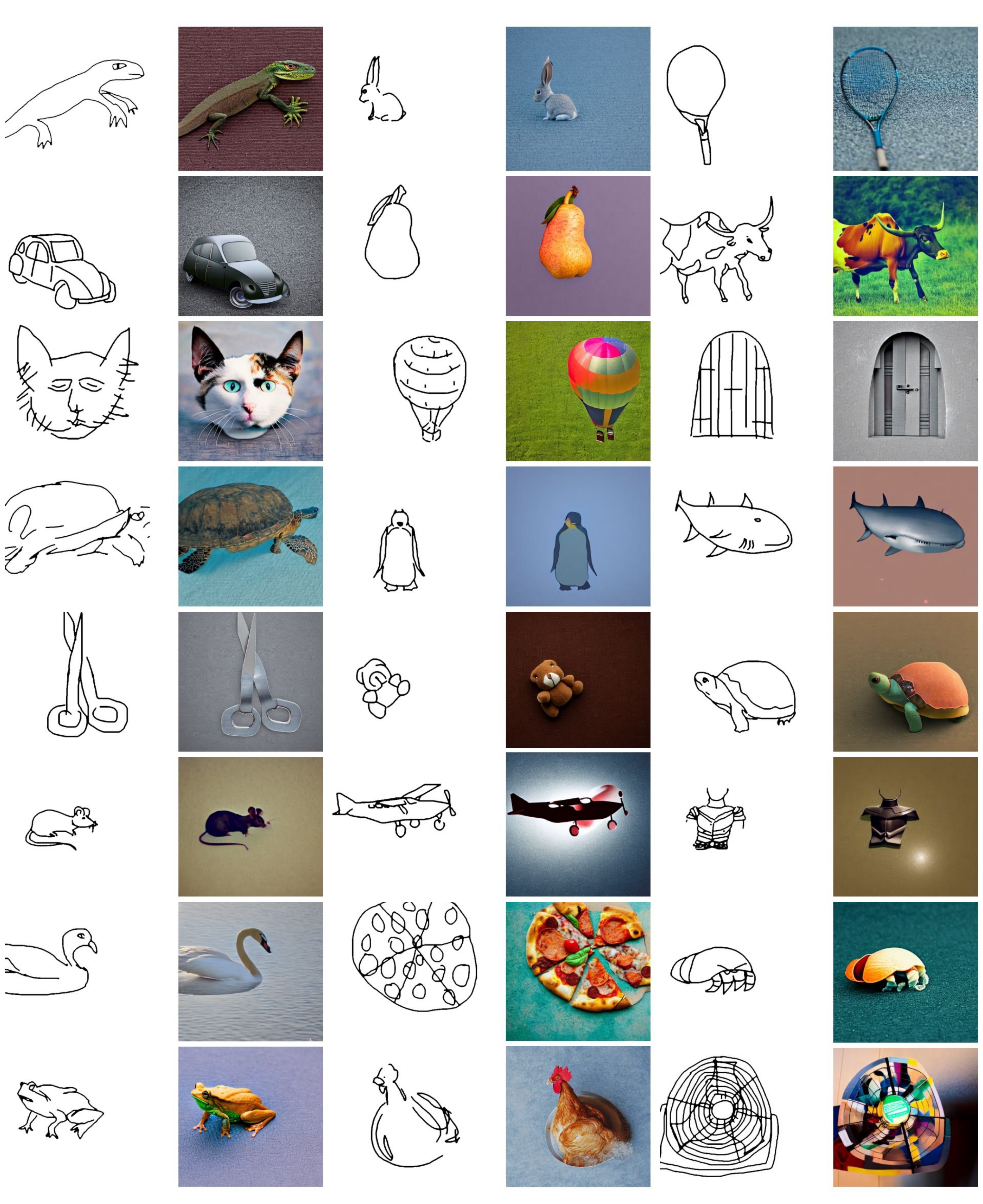}
    \caption{{\bf More Samples. } Samples of the proposed edge guidance being applied to Sketchy \cite{sangkloy2016sketchy} samples generated with the default guidance parameters. We use the classes names as prompts with the prefix "A photo of": "lizard", "rabbit", "racket", "car", "pear", "cow", "cat", "hot-air balloon", "door", "turtle", "penguin", "shark", "scissors", "teddy bear", "turtle", "mouse", "airplane", "armor", "swan", "pizza", "hermit crab", "frog", "chicken", "fan". Resampling, prompt-tuning, and guidance parameters tuning may significantly boost the visual quality of individual samples as well.}
    \label{fig:samples}
\end{figure*}

\begin{figure*}
    \centering
  \includegraphics[width=0.99\textwidth]{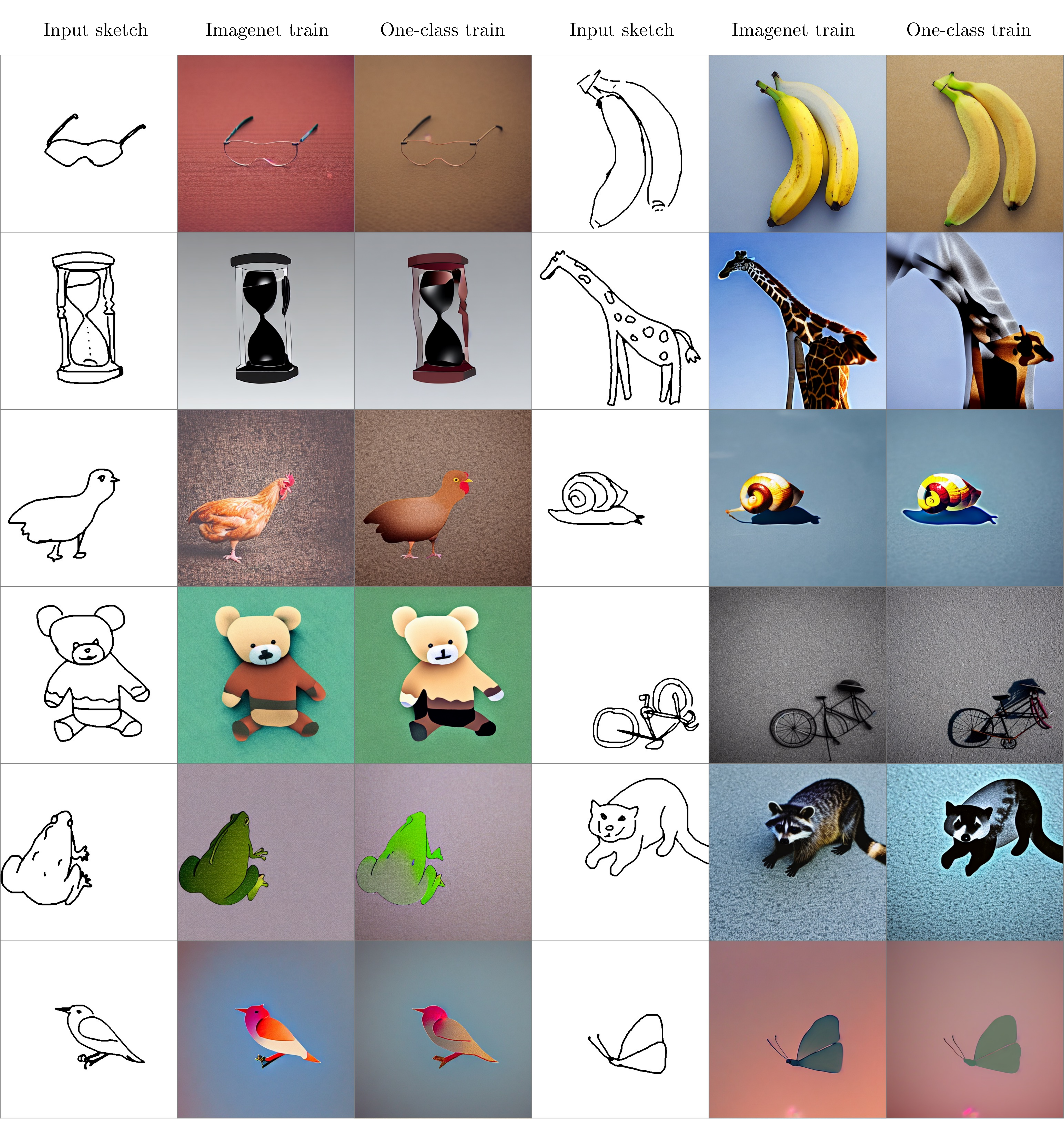}
    \caption{{\bf Training the LEP only on a single class vs. various classes} Comparison of guidance performed with a LEP model trained on various Imagenet samples, and only on images corresponding to different dog classes. Notably, due to the per-pixel nature of our training, even the model trained on a very specific image domain reasonably well generalizes to other domain samples. We use the classes names as prompts with the prefix "A photo of": "eyeglasses", "banana", "hourglass", "giraffe", "chicken" "snail", "teddy bear", "bicycle", "frog", "raccoon", "songbird", "butterfly", "cup", "fan".}
    \label{fig:limited_data}
\end{figure*}

\end{document}